\documentclass{article}

\usepackage[preprint]{neurips_2026}
\usepackage{graphicx}
\usepackage{caption}
\usepackage{subcaption}
\usepackage{multirow}
\usepackage{amsmath}
\usepackage{placeins}

\usepackage[utf8]{inputenc} 
\usepackage[T1]{fontenc}    
\usepackage{hyperref}       
\usepackage{url}            
\usepackage{booktabs}       
\usepackage{amsfonts}       
\usepackage{nicefrac}       
\usepackage{microtype}      
\usepackage{xcolor}         

\title{Observations Drift, Structures Remain: Structural Pretraining with Time Alignment for Electromagnetic Signals}

\author{%
Wenjin Gui \\
{\normalfont\small Beijing Institute of Technology}
\And
Junyu Shen \\
{\normalfont\small Tsinghua University}
\And
Haibo Xu \\
{\normalfont\small Hunan University}
\AND
Yuchuang Sun \\
{\normalfont\small Tsinghua University}
\And
Luqing Luo \\
{\normalfont\small Institute of Microelectronics} \\
{\normalfont\small Chinese Academy of Sciences}
\And
Wupeng Xie \\
{\normalfont\small AI Institute, CETC}
\And
Qian Huang \\
{\normalfont\small Tianjin University}
\AND
Fengxiang Wang\textsuperscript{*} \\
{\normalfont\small National University of Defense Technology}
\And
Yangang Sun\textsuperscript{*} \\
{\normalfont\small Tsinghua University}
\And
Maosong Sun\textsuperscript{*} \\
{\normalfont\small Tsinghua University}
}

\begin{document}

\maketitle

\begingroup
\renewcommand{\thefootnote}{*}
\footnotetext{Corresponding Author}
\endgroup

\begin{abstract}

Electromagnetic signal modeling is increasingly moving toward self-supervised pretraining on heterogeneous I/Q observations. However, an observed waveform is jointly shaped by signal generation, propagation environments, and acquisition processes, which mix reusable structural regularities with observation-specific variations. Existing methods usually pursue unified pretraining with reconstruction objectives on mixed observations. This heterogeneity creates two key mismatches for reconstruction pretraining: 
(1) appearance mismatch, induced by propagation-dependent waveform variations; and 
(2) temporal mismatch, induced by acquisition-dependent sampling coordinates.
These mismatches make observation-specific appearances and sampling coordinates easier to reconstruct than stable signal structures, inducing observational shortcuts. We identify this failure mode as observation-biased pretraining. To address this, we propose Structural Pretraining with Time Alignment for heterogeneous EM signals. It mitigates temporal mismatch with RoPE-fs, which maps token positions to sampling-rate-normalized physical-time positions, and reduces reliance on appearance shortcuts by replacing waveform reconstruction with masked prediction of discrete structural units. These units are organized into atomic structural units and emergent structural units that capture variable-length compositional patterns beyond fixed patch boundaries. Across diverse EM benchmarks, our method improves both full-supervision and few-shot adaptation over existing EM pretraining baselines, showing that the learned representations better preserve structural consistency when observations drift in waveform appearance and temporal scale.

\end{abstract}

\section{Introduction}

\begin{figure*}[t]
    \centering
    \includegraphics[width=0.98\textwidth]{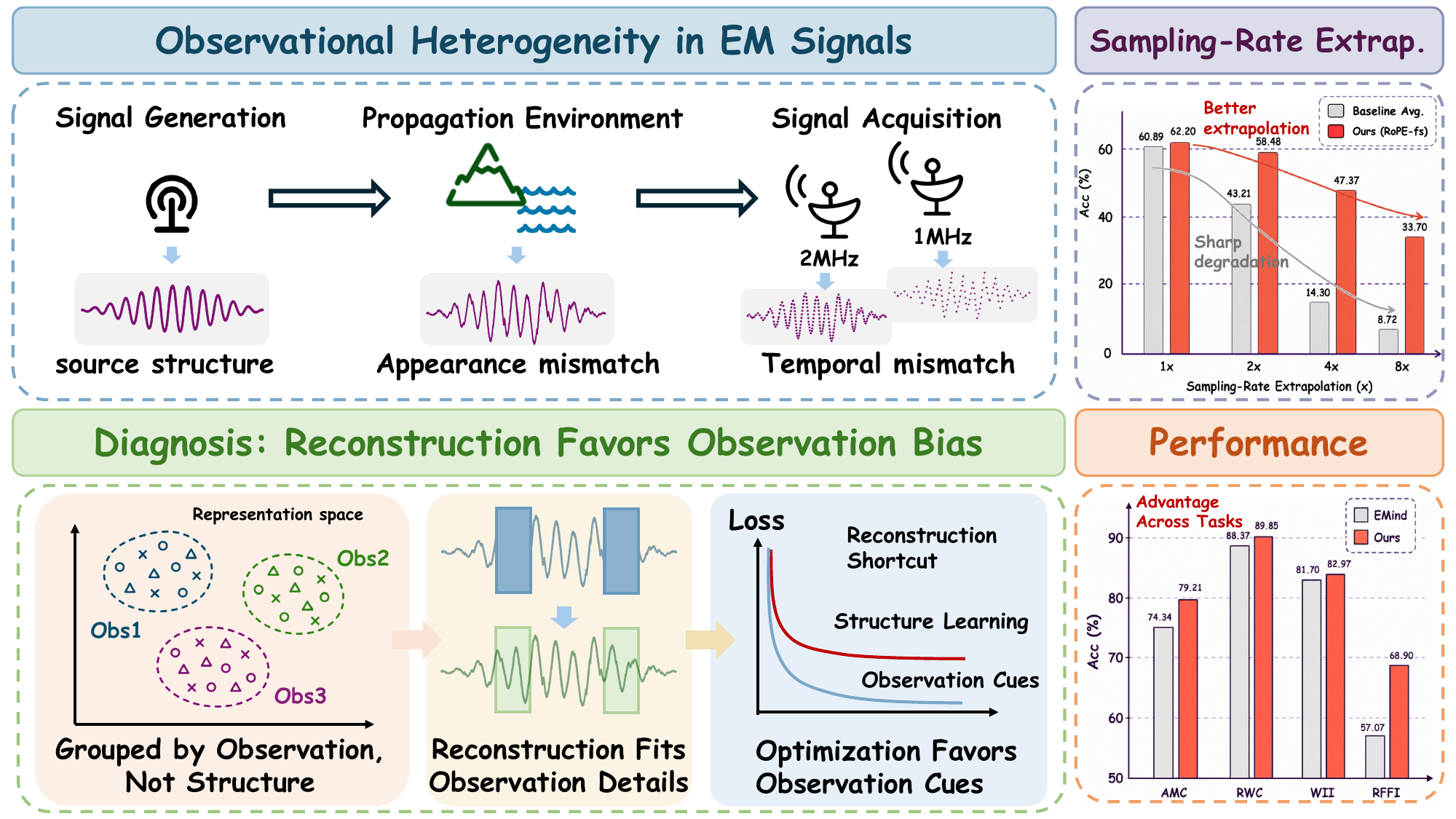}
    \caption{Motivation and diagnosis of observation-biased EM pretraining. Heterogeneous EM observations arise from source generation, propagation, and acquisition, causing appearance and temporal mismatches. Reconstruction objectives can easily exploit these observation-specific cues, leading representations to organize by observation conditions rather than reusable structures. The right panels summarize the resulting sampling-rate extrapolation degradation and the improved downstream performance achieved by our pretraining method.}
    \label{fig:intro}
\end{figure*}

Electromagnetic (EM) signal modeling is a fundamental problem for wireless communication, radar sensing, and spectrum monitoring. With the growing availability of multi-source I/Q data, recent studies have begun to move beyond task-specific architectures toward foundation-model-style pretraining on heterogeneous EM observations~\citep{chen2025radiollm, zhou2025spectrumfm, luo2025emind}. By learning from large-scale unlabeled signals and adapting the pretrained encoder to diverse downstream tasks, this paradigm aims to capture reusable structural regularities that are shared across signal types, datasets, and application scenarios. Therefore, a central goal of EM pretraining is to learn transferable representations from heterogeneous I/Q observations.

However, as summarized in Figure~\ref{fig:intro}, heterogeneous EM observations contain both reusable source structures and observation-specific variations introduced by propagation and acquisition. Our diagnostics reveal a counter-intuitive behavior of reconstruction-based pretraining on mixed I/Q observations: the representation space tends to cluster by observation conditions rather than stable structures. This weakens the structural semantics of the representation space, making it reflect observation-level similarity rather than reusable structural equivalence. We illustrate this mechanism as a reconstruction shortcut: when the target is the observed waveform, the model can reduce the loss by relying on observation-specific cues rather than reusable signal structures. This suggests that reconstruction-based pretraining over heterogeneous EM observations may not automatically produce structurally transferable representations. We refer to this failure mode as observation-biased pretraining. These findings raise the central question we aim to investigate:

\begin{center}
\fbox{
\begin{minipage}{0.95\linewidth}
\small\itshape
How can EM pretraining learn reusable structures in physical time from heterogeneous I/Q observations?
\end{minipage}
}
\end{center}

In response to this question, we identify two critical challenges that remain underexplored: 
(1) \textbf{temporal mismatch} induced by acquisition, and 
(2) \textbf{appearance mismatch} induced by propagation.
The former makes discrete token coordinates inconsistent with physical time across sampling rates, while the latter entangles reusable signal structures with observation-specific waveform appearances. 
Under reconstruction-based pretraining, both mismatches can be exploited as observational shortcuts, weakening the learning of cross-observation stable structures.

To address these challenges, we introduce \textbf{Structural Pretraining with Time Alignment}. 
For temporal mismatch, we align discrete patch positions with sampling-rate-normalized physical time through RoPE-fs, i.e., Physical-Time-Aware RoPE.
Instead of assigning the same positional meaning to a fixed token distance across all sampling rates, RoPE-fs represents token relations according to their physical-time intervals, making equivalent temporal structures more comparable under heterogeneous acquisition settings. 
For appearance mismatch, we move the learning target away from raw waveform reconstruction and instead perform masked prediction of structural-unit labels. 
Specifically, we construct a dictionary of atomic structural units from local I/Q patches and further induce emergent structural units through frequent adjacent-unit merging, allowing the pretraining target to describe recurring signal structures at multiple temporal scales. 
By coupling physical-time-aware positional modeling with atomic and emergent structural prediction, the framework encourages the encoder to focus on reusable signal regularities rather than observation-specific waveform details. Figure~\ref{fig:intro} also previews the advantages: our method mitigates sampling-rate extrapolation degradation, demonstrates a clear extrapolation advantage, and improves performance across diverse downstream EM tasks.

In summary, our main contributions are as follows:

(1) \textbf{Diagnosis.} We reveal a counter-intuitive failure mode in EM pretraining: reconstruction over mixed I/Q observations can favor observation-specific predictability over reusable structural regularities. We characterize this as observation-biased pretraining and trace it to temporal and appearance mismatches.

(2) \textbf{Framework.} We introduce \textit{structural pretraining with time alignment} for heterogeneous EM signals. The framework addresses temporal mismatch through RoPE-fs, and mitigates appearance-level reconstruction shortcuts through masked prediction of atomic and emergent structural units.

(3) \textbf{Evidence.} Across modulation, radar, interference, and fingerprinting benchmarks, our method improves both full-supervision and few-shot adaptation. Additional sampling-rate extrapolation, dictionary-quality, and waveform interpretation analyses further provide supporting evidence for the effectiveness of physical-time alignment and the learned structural units.

\section{Related Work}

\paragraph{Pretrained representation learning for EM signals.}
Inspired by the foundation model paradigm in language and vision~\citep{devlin2019bert, he2022masked}, recent studies have begun to move electromagnetic (EM) signal modeling beyond task-specific architectures toward pretrained representation learning on large-scale I/Q data. SpectrumFM and EMind demonstrate the transfer potential
of masked reconstruction or unified pretraining across communication and sensing tasks
\citep{zhou2025spectrumfm,luo2025emind}. These studies demonstrate the effectiveness of large-scale EM pretraining. However, existing methods typically optimize shared objectives directly on mixed observations~\citep{zhou2025spectrumfm, luo2025emind}. Since observed waveforms also depend on propagation and acquisition conditions, direct mixing can entangle reusable signal structures with observation-specific factors, making observation heterogeneity an under-addressed issue in EM pretraining.

\paragraph{Temporal granularity in time-series pretraining.}
Large-scale time-series pretraining has increasingly considered temporal granularity differences across domains and sensing systems. PatchTST and Moirai address this issue mainly through patch-wise or multi-scale tokenization~\citep{nie2022time, woo2024unified}, while MIRA extends rotary positional encoding to continuous timestamps so that attention can depend on real time intervals~\citep{li2025mira}. However, sampling-rate heterogeneity in EM signals presents a different setting: I/Q sequences are typically uniformly sampled within each source, but different sources may use different sampling rates, making the same token span correspond to different physical durations. This motivates a physical-time-aware positional design that aligns token relations with sampling-rate-dependent temporal semantics.

\paragraph{Self-supervised learning for continuous signal representation.}
Self-supervised learning for continuous signals, including speech, time series, and I/Q waveforms, commonly relies on reconstruction, contrastive prediction, or masked discrete prediction. Recent EM foundation models mainly adopt reconstruction objectives~\citep{zhou2025spectrumfm, luo2025emind}, while contrastive paradigms require carefully defined augmentations, positive pairs, and negatives under sampling-rate, channel, and environmental variations~\citep{oord2018representation, baevski2020wav2vec}. However, reconstructing heterogeneous EM observations may encourage the model to preserve observation-specific waveform details rather than reusable structures. Motivated by HuBERT, where discrete targets provide stable units for learning contextual regularities beyond low-level reconstruction~\citep{hsu2021hubert}, we move EM pretraining toward masked structural unit prediction with atomic and emergent units.


\section{Methods}

\subsection{Preliminaries}

We formalize a real electromagnetic observation as the result of three stages: source generation, environmental propagation, and acquisition. For the \(i\)-th signal, let \(z_i\) denote the latent source factor and \(g_i\) denote the generation mechanism. The clean source signal and the observed signal are formulated as
\begin{equation}
    s_i = \mathcal{G}_{g_i}(z_i),
    \qquad
    x_i = \mathcal{A}_{f_i}\big(\mathcal{E}_{\eta_i}(s_i)\big),
\end{equation}
where \(\mathcal{G}_{g_i}\) represents the source generation process, \(\mathcal{E}_{\eta_i}\) denotes environmental propagation under factors \(\eta_i\), and \(\mathcal{A}_{f_i}\) denotes acquisition with sampling rate \(f_i\).

This formulation separates source-induced generation heterogeneity, which defines the structural space to be learned, from observation-induced heterogeneity, which is introduced by propagation and acquisition and changes how source structures are presented to the model.

Observational heterogeneity induces two mismatches. First, sampling-rate differences introduce temporal mismatch. If the input is tokenized into patches of size \(P\), the physical-time coordinate of the \(l\)-th patch and the duration corresponding to a token distance \(\Delta l\) are
\begin{equation}
    \tau_{i,l}=\frac{(l-\frac{1}{2})P}{f_i},
    \qquad
    \Delta \tau_i=\frac{\Delta l \cdot P}{f_i}.
\end{equation}
Thus, under different sampling rates, the same physical structure may occupy different token spans, and the same discrete positional relation may carry different physical meanings. Second, environmental propagation introduces appearance mismatch: \(\mathcal{E}_{\eta_i}\) mixes source structures with environment-dependent factors, making the observed waveform appearance of \(x_i\) an unreliable proxy for the stable structure in \(s_i\).

\subsection{Observational Heterogeneity Biases Electromagnetic Pretraining}

\begin{figure*}[t]
    \centering
    \begin{subfigure}[t]{0.49\textwidth}
        \centering
        \includegraphics[width=\linewidth]{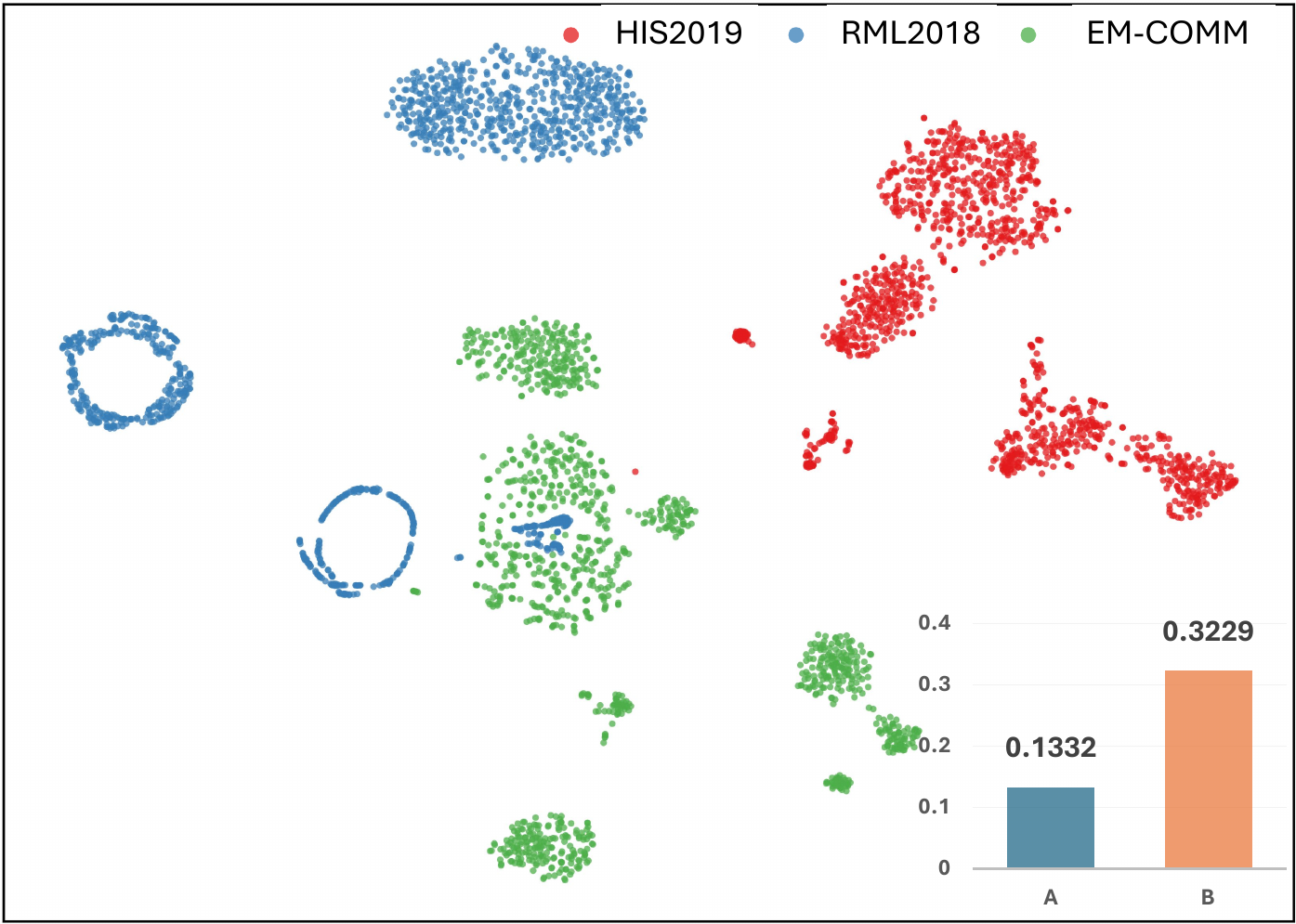}
        \caption{Cross-dataset heterogeneity, colored by dataset source.}
        \label{fig:vis_datasets_domain}
    \end{subfigure}
    \hfill
    \begin{subfigure}[t]{0.49\textwidth}
        \centering
        \includegraphics[width=\linewidth]{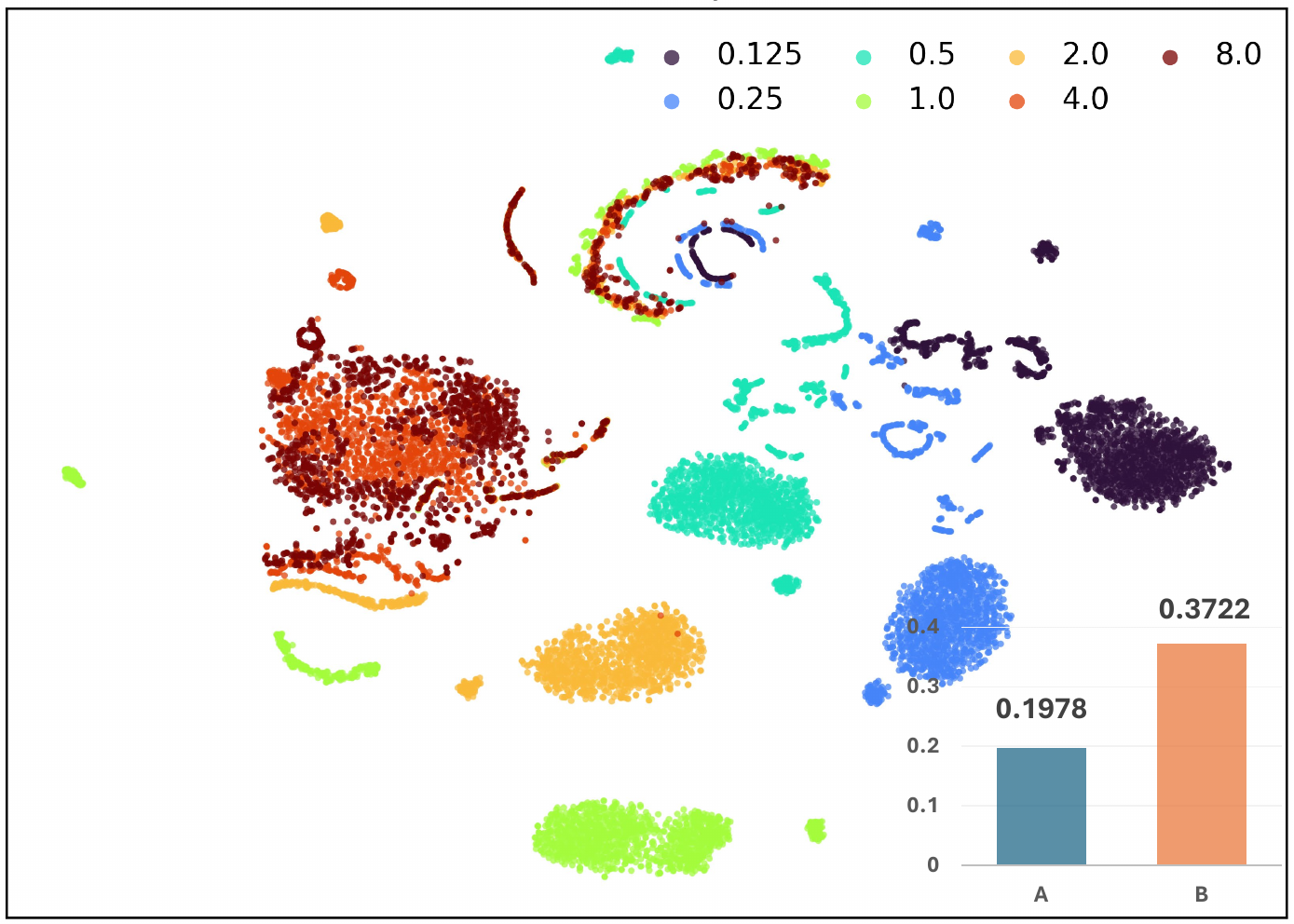}
        \caption{Sampling-rate heterogeneity on RML2018, colored by sampling-rate factor.}
        \label{fig:vis_rml2018_rate}
    \end{subfigure}

    \caption{
    Representation-space visualization under observational heterogeneity.
    Left: cross-dataset heterogeneity.
    Right: sampling-rate heterogeneity.
    Statistics report cosine similarities for A: same structure, different observations; B: different structures, same observation.
    Modulation-colored visualizations are provided in Figure~\ref{fig:heterogeneity_visualization_appendix}.
    }
    \label{fig:heterogeneity_visualization}
\end{figure*}



A desirable EM representation should preserve structural equivalence across observation conditions: same-structure signals should remain close across datasets or sampling rates, while different structures should stay separable within the same condition. Figure~\ref{fig:heterogeneity_visualization} visualizes the pretrained representations of EMind across datasets and sampling rates, revealing a clear observation-biased pattern. In both settings, embeddings cluster predominantly by observation conditions rather than modulation structures.

We further quantify this pattern using cosine similarity and linear probing on EMind and SpectrumFM representations. In terms of cosine similarity, structure-preserving representations should satisfy A (same structure, different observations) > B (different structures, same observation), yet both models exhibit A < B across observation conditions. Linear-probing results further show that dataset identity is highly separable in the pretrained representations, with higher prediction accuracy than modulation identity, as detailed in Appendix~\ref{app:pre_analysis}. This indicates that reconstruction-pretrained representations are fragmented by dataset source or sampling-rate factor, weakening structural equivalence across heterogeneous observations.

\begin{figure*}[t]
    \centering
    \includegraphics[width=0.98\textwidth]{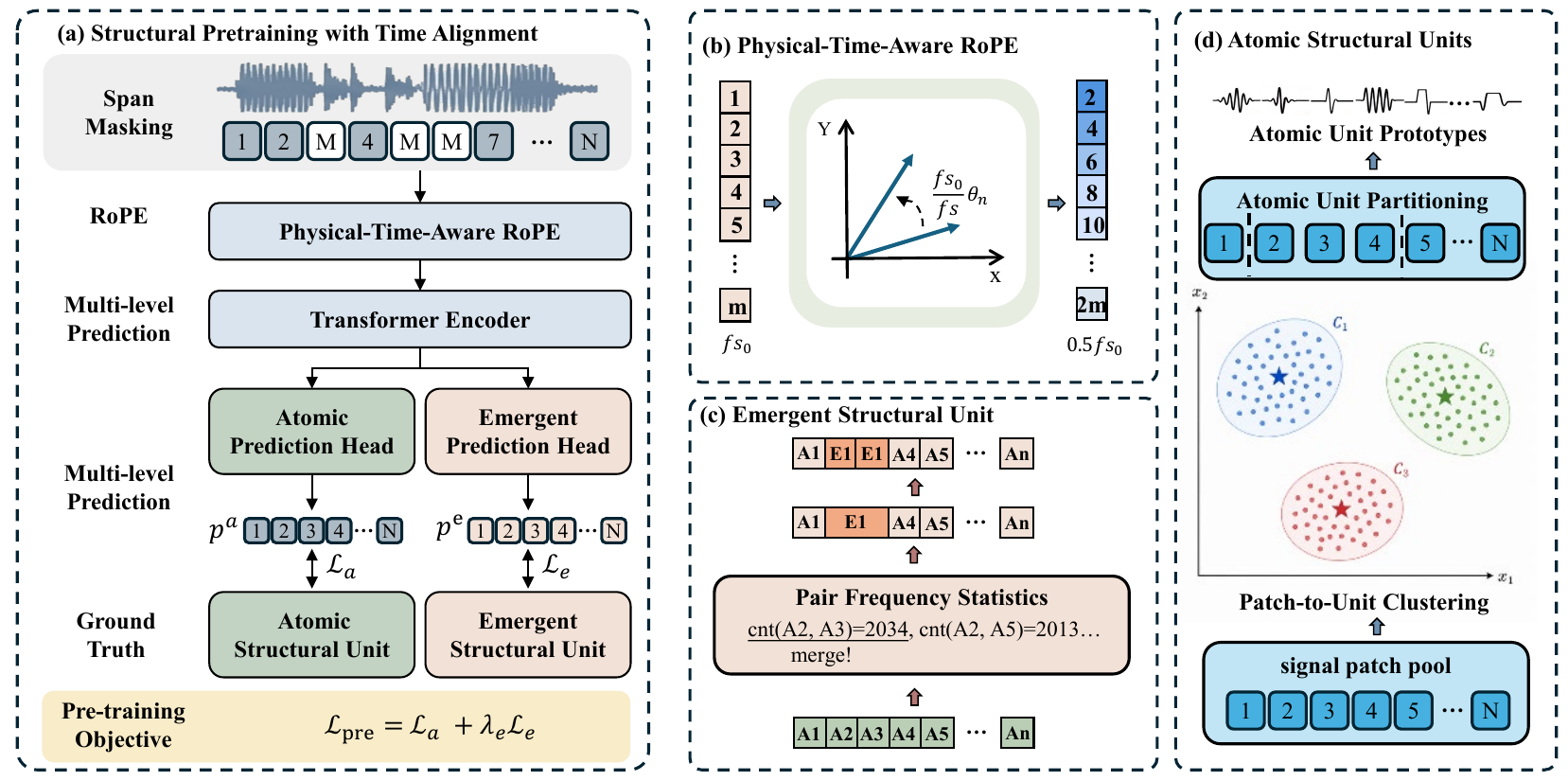}
    \caption{Overview of structural pretraining with time alignment. I/Q patches are encoded with RoPE-fs, while masked prediction of atomic and emergent structural-unit labels provides multi-level structural supervision instead of raw waveform reconstruction.}
    \label{fig:framework}
\end{figure*}

This bias is inherent to objectives defined on final observations. A masked reconstruction loss,
\begin{equation}
    \mathcal{L}_{\mathrm{rec}}=\sum_{n\in\Omega}\ell(\hat{x}[n],x[n]),
\end{equation}
rewards accurate prediction of the observed waveform, without distinguishing stable source structures from observation-induced regularities. Under heterogeneous observations, sampling-grid-dependent temporal patterns and channel-, noise-, or data-source-specific appearances may therefore become shortcuts for reducing the loss. This motivates our design: RoPE-fs mitigates temporal mismatch, while masked structural-unit prediction reduces reliance on observation-specific waveform appearance. Figure~\ref{fig:framework} provides an overview of the proposed framework, which couples RoPE-fs signal encoding with masked prediction of atomic and emergent structural units.

\subsection{RoPE-fs: Physical-Time-Aware Signal Encoding}

As illustrated in Figure~\ref{fig:framework}, to mitigate temporal mismatch, RoPE-fs encodes signal patches using sampling-rate-normalized physical-time positions rather than raw discrete indices.

\paragraph{I/Q patch tokenization.}
Given an observed I/Q signal \(x_i\in\mathbb{R}^{2\times T_i}\), we divide it into non-overlapping temporal patches with patch size \(P\):
\begin{equation}
    x_i=[p_{i,1},\ldots,p_{i,L_i}],\quad
    p_{i,l}\in\mathbb{R}^{2\times P},\quad
    L_i=\lfloor T_i/P\rfloor .
\end{equation}
Each patch is projected into a \(d\)-dimensional embedding by a shared patch embedding function \(\phi(\cdot)\):
\begin{equation}
    e_{i,l}=\phi(p_{i,l}),\qquad
    E_i=[e_{i,1},\ldots,e_{i,L_i}]\in\mathbb{R}^{L_i\times d}.
\end{equation}

\paragraph{RoPE-fs.}
Standard positional encodings use the discrete token index \(l\), whose physical meaning varies with the sampling rate in EM signals. We therefore replace \(l\) with a sampling-rate-normalized physical-time position. Let \(f_0\) denote a reference sampling rate. Following the physical-time coordinate \(\tau_{i,l}\) defined above, we define
\begin{equation}
    \tilde{l}_{i,l}
    =
    \frac{\tau_{i,l} f_0}{P}
    =
    \left(l-\frac{1}{2}\right)\frac{f_0}{f_i},
    \qquad
    \theta_{i,l,m}=\tilde{l}_{i,l}\omega_m .
\end{equation}
We then apply rotary positional encoding~\citep{su2024roformer} using \(\tilde{l}_{i,l}\) instead of the discrete index \(l\), where \(\theta_{i,l,m}\) is the rotation angle for the \(m\)-th rotary frequency \(\omega_m\). Thus, positional phases depend on normalized physical time, assigning more consistent phases to signal segments occupying similar physical-time intervals across sampling rates. The resulting token sequence is processed by a Transformer encoder~\citep{vaswani2017attention}, yielding contextual representations \(H_i=[h_{i,1},\ldots,h_{i,L_i}]\), where \(h_{i,l}\in\mathbb{R}^{d}\).

\subsection{Structural Unit Discovery}

To mitigate appearance bias, we discover discrete structural units from patch-level signal representations and use them as self-supervised targets, since EM signals do not provide native symbols like natural language. The dictionary is not intended to remove all environment-related information, but to suppress unstable sample-specific details and preserve recurring corpus-level structures.

\paragraph{Atomic structural units.}
Given a pretraining corpus \(\mathcal{D}\), we associate each patch position with a structural representation \(r_{i,l}\in\mathbb{R}^{d_r}\) and collect all patch-level representations:
\begin{equation}
    \mathcal{R}=\{r_{i,l}\mid x_i\in\mathcal{D},\;1\leq l\leq L_i\}.
\end{equation}
\(r_{i,l}\) can be initialized from raw patch features and iteratively refined by re-extracting encoder representations after structural pretraining.
We learn an atomic codebook \(\mathcal{C}^{a}=\{c_k^{a}\}_{k=1}^{K_a}\) by clustering \(\mathcal{R}\), and assign each patch to its nearest centroid:
\begin{equation}
\begin{aligned}
    \mathcal{C}^{a}
    &= \arg\min_{\{c_k^a\}_{k=1}^{K_a}}
    \sum_{r\in\mathcal{R}}\min_{1\leq k\leq K_a}\|r-c_k^a\|_2^2,\\
    y_{i,l}^{a}
    &= \arg\min_{1\leq k\leq K_a}\|r_{i,l}-c_k^a\|_2^2 .
\end{aligned}
\end{equation}
This yields an atomic structural sequence \(Y_i^a=[y_{i,1}^a,\ldots,y_{i,L_i}^a]\), where \(y_{i,l}^a\in\{1,\ldots,K_a\}\). Atomic units provide dense patch-level structural anchors and define the base alphabet for higher-order structures.

\paragraph{Emergent structural units.}
Atomic units describe local structures at a fixed patch resolution, while stable EM patterns may span multiple adjacent patches. Relying only on isolated atomic labels can therefore fragment recurring variable-span structures. To capture such patterns, we induce \emph{emergent structural units} (ESUs) from atomic sequences. Each ESU \(e\) expands to an atomic subsequence \(e\Rightarrow[a_1,\ldots,a_m]\), where \(a_j\in V^a\) and \(m\) is its span length.

Inspired by byte-pair encoding~\citep{gage1994new}, we start from atomic label sequences and iteratively merge the most recurrent adjacent pair \((u,v)\), where each unit may be atomic or previously merged. In this work, the merge rule is
\begin{equation}
    (u^{*},v^{*})=
    \arg\max_{(u,v):\,m(u)+m(v)\leq m_{\max}}
    \mathrm{count}_{\mathcal{D}}(u,v).
\end{equation}
The span constraint prevents uncontrolled unit growth. Repeating this process yields $K_e$ merged ESUs. Together with the $K_a$ atomic singleton units, they form the ESU vocabulary $V^e = \{1,\ldots,K_a+K_e\}$. Applying the learned merge rules to \(Y_i^a\) produces a compressed ESU sequence
\begin{equation}
    \bar{Y}_i^e=[\bar{y}_{i,1}^e,\ldots,\bar{y}_{i,M_i}^e],
    \qquad M_i\leq L_i,
\end{equation}
where each element belongs to \(V^e\) and may cover one or more atomic positions.

\paragraph{Patch-aligned structural labels.}
Since ESU sequences are variable-length but encoder outputs are patch-aligned, we backfill each ESU identity to its covered patch positions as token-level targets. Specifically, if an ESU \(e\) covers positions \(s,\ldots,s+m-1\), we set \(y_{i,t}^{e}=e\) for \(t=s,\ldots,s+m-1\), yielding a patch-aligned sequence \(Y_i^e=[y_{i,1}^e,\ldots,y_{i,L_i}^e]\). Each patch then has both an atomic label \(y_{i,l}^a\in V^a\) and an emergent label \(y_{i,l}^e\in V^e\), which serve as local and higher-order targets for masked structural prediction.

\subsection{Masked Multi-level Structural Unit Prediction}

The preceding sections define two prerequisites for structural pretraining: physical-time-aware representations and patch-aligned structural labels. We now combine them into a masked prediction objective. Unlike waveform reconstruction, which asks the model to reproduce observation-specific continuous values, our objective trains the encoder to recover discrete structural identities from masked I/Q contexts. This shifts pretraining from observation-level predictability to structure-level predictability.

\paragraph{Span masking on I/Q signals.}
Given an input signal \(x_i\) with \(L_i\) patch positions, we sample masked positions \(\Omega_i\subseteq\{1,\ldots,L_i\}\). Following span-masking practice~\citep{hsu2021hubert}, \(\Omega_i\) consists of consecutive patch spans to reduce trivial local interpolation. The mask is applied to the corresponding continuous I/Q patches, producing a corrupted signal \(\tilde{x}_i\), which is encoded as
\begin{equation}
    \tilde{H}_i=F_{\theta}(\tilde{x}_i,f_i)
    =[\tilde{h}_{i,1},\ldots,\tilde{h}_{i,L_i}].
\end{equation}
Losses are computed only for \(l\in\Omega_i\), forcing the model to infer structural identities from context.

\paragraph{Atomic and emergent structural prediction.}
For each masked position, two lightweight heads predict the patch-aligned atomic and emergent labels \(y_{i,l}^{a}\) and \(y_{i,l}^{e}\):
\begin{equation}
    p_{i,l}^{a}=\mathrm{softmax}(W_a\tilde{h}_{i,l}),
    \qquad
    p_{i,l}^{e}=\mathrm{softmax}(W_e\tilde{h}_{i,l}).
\end{equation}
Here $W_a$ and $W_e$ classify over $V^a$ and $V^e$, respectively, where $|V^a|=K_a$ and $|V^e|=K_a+K_e$. Atomic prediction preserves fine-grained local discriminability, while emergent prediction encourages the encoder to recover higher-order recurrent compositions beyond fixed patch boundaries.

\paragraph{Joint objective.}
The atomic and emergent prediction losses are computed over masked positions and combined as
\begin{equation}
    \mathcal{L}_{\mathrm{pre}}
    =
    \mathcal{L}_{a}
    +
    \lambda_e \mathcal{L}_{e},
\end{equation}
where \(\mathcal{L}_{a}\) and \(\mathcal{L}_{e}\) are cross-entropy losses for atomic and emergent prediction on masked positions, and \(\lambda_e\) balances emergent supervision.

Overall, masked multi-level structural prediction completes the transition from observation-biased waveform modeling to physical-time-aware structural pretraining. The atomic level grounds the model in local waveform primitives, while the emergent level guides it toward longer and more stable compositional structures.

\section{Experiments}


Our experiments examine three aspects: downstream adaptation under sufficient and limited supervision, component contributions through ablations, and whether RoPE-fs and the learned dictionary reduce temporal mismatch and form stable structural units.

\subsection{Experimental Setup}

\paragraph{Datasets and Tasks.}
We construct the unlabeled pretraining corpus from the dataset suite proposed by EMind~\citep{luo2025emind}, with details provided in Appendix. To avoid data leakage, the corpus is disjoint from downstream validation and test splits, and no downstream validation or test samples are used for dictionary construction, emergent-unit mining, or pretraining. Downstream benchmarks include RML2016.10A, RML2016.04C, RML2016.10B, RML2018.01A~\citep{o2016radio, o2018over}, Panoradio~\citep{scholl2019classification}, RadChar~\citep{huang2023multi}, EM-Infer-Comm, EM-Infer-Radar, and EM-AIS~\citep{luo2025emind}, covering modulation, radar, interference, and emitter-identification tasks. Few-shot evaluation is conducted on modulation benchmarks.

\paragraph{Compared Methods.} We compare with two types of baselines. First, ResNet~\citep{he2016deep} and Transformer~\citep{vaswani2017attention} are used as representative supervised I/Q classifiers. We focus on these general architectures instead of exhaustive dataset-specific EM models because prior EM foundation models, such as EMind, have already shown that large-scale pretraining can surpass many task-specific designs; they thus provide simple and consistent supervised references across tasks. Second, we compare with pretrained EM foundation models, including SpectrumFM~\citep{zhou2025spectrumfm} and EMind~\citep{luo2025emind}, as stronger self-supervised baselines.

\paragraph{Model and Training Configuration.}
We use a 12-layer 1D Transformer encoder with hidden size 256, 8 heads, patch size 8, and 99th-percentile amplitude normalization. The atomic vocabulary contains $K_a = 512$ units. For emergent structural prediction, we perform $K_e = 512$ merges; the resulting ESU vocabulary includes both atomic singleton units and merged units, yielding $K_a + K_e = 1024$ prediction targets, with expansion length up to 5. Pretraining uses span masking with probability 0.06 and span length 8. Downstream prediction is performed by  mean pooling final-layer patch representations and feeding them into a task-specific head. Following HuBERT-style refinement~\citep{hsu2021hubert}, Iter0 builds dictionaries from raw patch features, and Iter1 rebuilds them from pretrained representations for a second pretraining round. Details are in Appendix.


\subsection{Main Results}

\begin{table*}[t]
\centering
\caption{Full-supervision fine-tuning results on downstream EM benchmarks.}
\label{tab:full_supervision}

\resizebox{\textwidth}{!}{%
\begin{tabular}{lccccccccc}
\toprule
\textbf{Task}
& \multicolumn{5}{c}{\textbf{AMC}}
& \multicolumn{1}{c}{\textbf{RWC}}
& \multicolumn{2}{c}{\textbf{WII}}
& \multicolumn{1}{c}{\textbf{RFFI}} \\
\cmidrule(lr){2-6}
\cmidrule(lr){7-7}
\cmidrule(lr){8-9}
\cmidrule(lr){10-10}
\textbf{Method}
& \textbf{RML16A}
& \textbf{RML16B}
& \textbf{RML16C}
& \textbf{RML18A}
& \textbf{Panoradio}
& \textbf{RadChar}
& \textbf{EI-Comm}
& \textbf{EI-Radar}
& \textbf{EM-AIS} \\
\midrule
ResNet
& 50.04 & 54.88 & 55.97 & 43.06 & 71.18 & 87.61 & 76.79 & 71.64 & 42.02 \\
Transformer
& 59.27 & 63.10 & 65.41 & 59.45 & 84.93 & 88.77 & 80.44 & 77.05 & 39.12 \\
SpectrumFM
& 63.72 & 65.35 & 73.37 & -- & -- & -- & -- & -- & -- \\
EMind
& 62.51 & 65.45 & 74.34 & 63.83 & 81.97 & 88.37 & 81.70 & 79.19 & 57.07 \\
\midrule
Ours
& \textbf{64.07}{\scriptsize~(+1.56)}
& \textbf{66.23}{\scriptsize~(+0.78)}
& \textbf{79.21}{\scriptsize~(+4.87)}
& \textbf{63.91}{\scriptsize~(+0.08)}
& \textbf{95.13}{\scriptsize~(+13.16)}
& \textbf{89.85}{\scriptsize~(+1.48)}
& \textbf{82.97}{\scriptsize~(+1.27)}
& \textbf{79.87}{\scriptsize~(+0.68)}
& \textbf{68.90}{\scriptsize~(+11.83)} \\
\bottomrule
\end{tabular}%
}

\vspace{2pt}
\begin{minipage}{0.98\textwidth}
\footnotesize
\textit{Note.} EI-Comm and EI-Radar denote EM-Infer-Comm and EM-Infer-Radar, respectively. 
``--'' indicates unavailable SpectrumFM results; the corresponding pretrained weights and pretraining data were not publicly available to us for reliable reproduction on these benchmarks.
\end{minipage}
\end{table*}

\paragraph{Full Supervision.}

Following EMind~\citep{luo2025emind}, we group downstream benchmarks into Automatic Modulation Classification (AMC), Radar Waveform Classification (RWC), Wireless Interference Identification (WII), and Radio Frequency Fingerprint Identification (RFFI). Table~\ref{tab:full_supervision} reports full-supervision fine-tuning results across nine benchmarks. Our method achieves the best performance on all datasets. Compared with supervised ResNet and Transformer baselines, the results show the advantage of pretrained EM representations for downstream adaptation. More importantly, our method consistently improves over pretrained EM foundation models such as SpectrumFM and EMind, suggesting that masked structural-unit prediction is a more effective pretraining objective than directly modeling continuous waveform appearance. These results indicate that the learned representations generalize beyond standard modulation recognition and adapt effectively to broader downstream EM tasks such as interference recognition and emitter identification.


\paragraph{Few-shot.}

\begin{table*}[t]
\centering
\caption{Few-shot fine-tuning results on AMC benchmarks.}
\label{tab:few_shot}

\resizebox{\textwidth}{!}{%
\begin{tabular}{lcccccccc}
\toprule
\textbf{Dataset}
& \multicolumn{2}{c}{\textbf{RML16A}}
& \multicolumn{2}{c}{\textbf{RML16B}}
& \multicolumn{2}{c}{\textbf{RML16C}}
& \multicolumn{2}{c}{\textbf{RML18A}} \\
\cmidrule(lr){2-3}
\cmidrule(lr){4-5}
\cmidrule(lr){6-7}
\cmidrule(lr){8-9}
\textbf{Method}
& \textbf{50-shot} & \textbf{100-shot}
& \textbf{50-shot} & \textbf{100-shot}
& \textbf{50-shot} & \textbf{100-shot}
& \textbf{50-shot} & \textbf{100-shot} \\
\midrule
ResNet
& 39.26 & 41.55
& 42.42 & 43.56
& 44.08 & 46.50
& 31.23 & 32.24 \\
Transformer
& 44.08 & 45.70
& 46.19 & 49.22
& 48.41 & 50.43
& 35.04 & 38.59 \\
EMind
& 48.38 & 50.10
& 49.08 & 51.20
& 55.01 & 57.09
& 41.97 & 49.13 \\
\midrule
Ours
& \textbf{50.76}{\scriptsize~(+2.38)}
& \textbf{52.40}{\scriptsize~(+2.30)}
& \textbf{50.26}{\scriptsize~(+1.18)}
& \textbf{53.31}{\scriptsize~(+2.11)}
& \textbf{57.27}{\scriptsize~(+2.26)}
& \textbf{60.42}{\scriptsize~(+3.33)}
& \textbf{45.97}{\scriptsize~(+4.00)}
& \textbf{52.31}{\scriptsize~(+3.18)} \\
\bottomrule
\end{tabular}%
}
\end{table*}

To further evaluate transferability under limited supervision, we follow the few-shot setting in EMind~\citep{luo2025emind} and fine-tune on AMC benchmarks with 50 or 100 labeled examples for each class-SNR pair. As shown in Table~\ref{tab:few_shot}, our method achieves the best performance in all eight settings. Compared with EMind, our method obtains consistent gains ranging from +1.18 to +4.00 points, with particularly clear improvements on RML18A. These results indicate that the proposed pretraining does not merely improve full-data adaptation, but also learns reusable structural representations that can be effectively transferred with limited labeled data.

\subsection{Ablation Study}



Table~\ref{tab:component_ablation} examines the contributions of sampling-rate-aware encoding and structural-unit modeling. Comparing \emph{abs+fix} with \emph{RoPE-fs+fix}, RoPE-fs brings clear gains in several settings, suggesting that injecting sampling-rate information into positional encoding helps reduce temporal-coordinate mismatch. However, its benefit is not always stable when the target remains fixed patch-level labels, indicating that sampling-rate awareness alone cannot fully address the structural granularity mismatch in EM signals. In contrast, combining RoPE-fs with atomic and emergent units supervision gives the strongest overall results. Notably, under this structured label space, iter1 consistently improves over iter0 across all datasets, while fixed-label variants show inconsistent iteration behavior. This suggests that iterative dictionary refinement becomes effective only when pseudo labels are organized into stable multi-level structural units, supporting our design of coupling physical-time alignment with structural abstraction.

Additional controlled ablations in Table~\ref{tab:additional_component_ablations} further confirm the benefits of multi-level structural supervision and RoPE-fs, while a compute-matched comparison shows that iterative dictionary refinement outperforms continued training with fixed dictionaries. Hyperparameter sensitivity results in Table~\ref{tab:hyperparameter_sensitivity} show that the adopted configurations achieve the highest accuracy among the tested settings on both evaluated datasets.

\subsection{Further Analyses}
\paragraph{Sampling-rate Extrapolation.}

\begin{table*}[t]
\centering
\caption{Ablation and sampling-rate extrapolation analyses.}
\label{tab:ablation_extrapolation}
\footnotesize

\begin{subtable}[t]{0.46\textwidth}
\centering
\caption{Component-wise ablation}
\label{tab:component_ablation}
\resizebox{\linewidth}{!}{%
\begin{tabular}{lcccc}
\toprule
\textbf{Data} & \textbf{Iter.} & \textbf{Abs.+Fix} & \textbf{R-fs+Fix} & \textbf{R-fs+MU} \\
\midrule
RML16C & 0 & 65.71 & 77.23 & \textbf{78.37} \\
RML16C & 1 & 66.24 & 74.93 & \textbf{79.21} \\
RML16A & 0 & 62.15 & \textbf{63.16} & 62.04 \\
RML16A & 1 & 62.53 & 62.78 & \textbf{64.07} \\
RML16B & 0 & \textbf{65.33} & 65.00 & 64.66 \\
RML16B & 1 & 65.41 & 64.75 & \textbf{66.23} \\
RML18A & 0 & 60.77 & 61.96 & \textbf{62.48} \\
RML18A & 1 & 61.40 & 62.02 & \textbf{63.91} \\
\bottomrule
\end{tabular}%
}
\end{subtable}
\hspace{0.015\textwidth}
\begin{subtable}[t]{0.48\textwidth}
\centering
\caption{Sampling-rate extrapolation}
\label{tab:sr_extrapolation}
\resizebox{\linewidth}{!}{%
\begin{tabular}{llcccc}
\toprule
\textbf{Data} & \textbf{Split} & \textbf{Abs.-S} & \textbf{Abs.-M} & \textbf{RoPE-M} & \textbf{RoPE-fs-M} \\
\midrule
RML16C & Seen   & 62.67 & 61.38 & 62.65 & \textbf{62.78} \\
       & Unseen & 24.43 & 33.68 & 32.60 & \textbf{40.26} \\
RML16A & Seen   & 58.47 & 56.94 & 58.32 & \textbf{59.22} \\
       & Unseen & 23.58 & 31.01 & 31.70 & \textbf{37.86} \\
RML16B & Seen   & \textbf{64.34} & 62.30 & 62.05 & 61.96 \\
       & Unseen & 23.55 & 33.31 & 31.19 & \textbf{39.30} \\
RML18A & Seen   & \textbf{61.57} & 57.72 & 59.80 & 60.27 \\
       & Unseen & 16.19 & 23.12 & 22.46 & \textbf{33.64} \\
\bottomrule
\end{tabular}%
}
\end{subtable}

\vspace{2pt}
\begin{minipage}{0.96\textwidth}
\scriptsize
\textit{Note.} R-fs denotes RoPE-fs. MU denotes multi-level structural units. 
Abs.-S and Abs.-M denote single-rate and multi-rate training with absolute positional encoding; 
RoPE-M and RoPE-fs-M denote multi-rate training with standard RoPE and RoPE-fs, respectively.
\end{minipage}
\end{table*}

Table~\ref{tab:sr_extrapolation} evaluates sampling-rate extrapolation, where seen and unseen rates are constructed by resampling signals with different factors. \emph{Abs.-Single} suffers severe degradation on unseen rates, showing that standard encoders do not naturally generalize across sampling-rate shifts. Multi-rate resampling improves \emph{Abs.-Multi}, but the remaining gap and the inconsistent gains of \emph{RoPE-Multi} suggest that augmentation and index-based relative positions are insufficient. \emph{RoPE-fs-Multi} achieves the best unseen performance on all four datasets, with Figure~\ref{fig:sampling_rate_factor} further showing consistent gains across unseen factors. This supports the role of physical-time-aware positional modeling.

\paragraph{Dictionary Quality.}

Table~\ref{tab:atomic_dict_quality} shows that the atomic dictionary maintains diverse code usage and non-trivial temporal transitions, while its moderate NMI with mode and dataset identity suggests local structural targets rather than global-label proxies. Table~\ref{tab:bpe_dict_quality} further indicates that BPE merging activates the emergent vocabulary effectively and captures reusable variable-length compositions, as reflected by high vocabulary usage, increased token length, reduced compression ratio, and non-negligible long-unit usage. These results support the learned dictionaries as compact structural supervision for pretraining. We further evaluate the stability of learned representations and structural units under signal perturbations. As shown in Table~\ref{tab:structural_unit_stability}, pretraining improves representation robustness to diverse perturbations, while both atomic and emergent structural units exhibit greater cross-perturbation consistency.

\begin{figure*}[t]
    \centering
    \includegraphics[width=\textwidth]{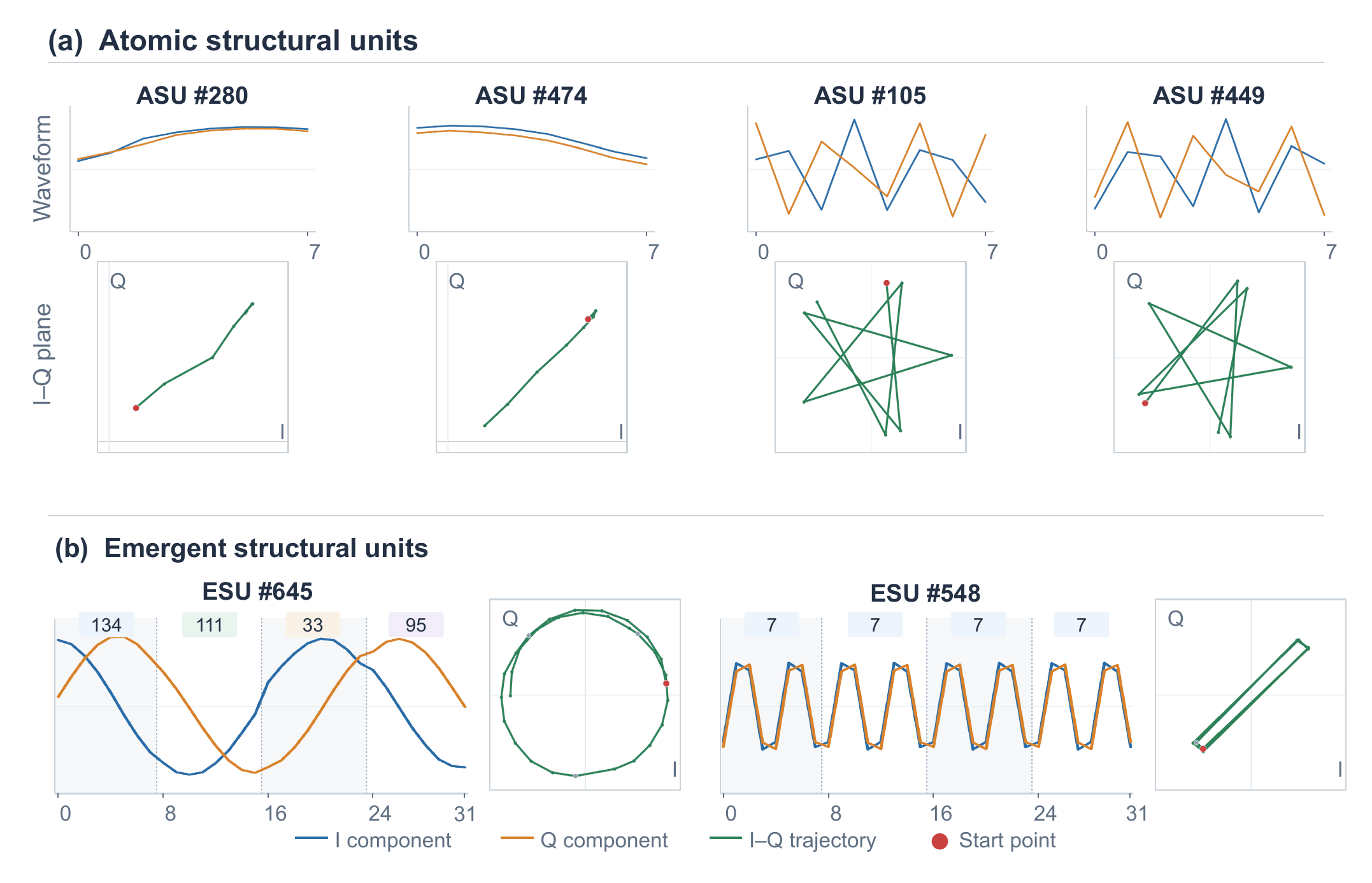}
    \caption{
    Waveform patterns of learned structural units.
    (a) Representative atomic structural units (ASUs).
    (b) Representative emergent structural units (ESUs) and their constituent atomic sequences.
    }
    \label{fig:structural_prototypes}
\end{figure*}

\paragraph{Waveform Interpretation of Structural Units.}
To examine the waveform structures captured by the learned units, we visualize Iter0 ASU cluster centers and construct ESU templates by concatenating their constituent ASU prototypes. Figure~\ref{fig:structural_prototypes} visualizes representative units in the time domain and I--Q plane.

ASUs capture distinct local waveform primitives. Tokens \#280 and \#474 represent rising and falling amplitude-envelope edges, exhibiting outward and inward I--Q trajectories, respectively. Tokens \#105 and \#449 capture rapid local phase rotations, characterized by regular oscillations and star-like I--Q trajectories. At a higher level, ESUs encode longer compositional patterns. ESU \#645 connects successive phase fragments into a near-complete rotational cycle, while ESU \#548 extends a periodic pattern through repeated atomic units. These observations demonstrate that ASUs capture interpretable local structures, while ESUs further model their sequential composition and repetition over longer temporal spans.

\section{Conclusion}

Electromagnetic pretraining over heterogeneous I/Q observations should capture reusable signal structures and physical-time relations rather than observation-specific waveform appearances. We show that reconstruction on mixed observations can induce observation-biased representations, where sampling-rate-dependent coordinates and propagation-induced appearances dominate stable structural regularities. We therefore introduce structural pretraining with time alignment that aligns patch positions with RoPE-fs and replaces waveform reconstruction with masked prediction of atomic and emergent structural units. Across modulation, radar, interference, and fingerprinting benchmarks, our method improves full-supervision and few-shot adaptation, while further analyses support physical-time alignment and reveal interpretable local waveform primitives and their longer compositional patterns. These results highlight the importance of organizing EM self-supervision around stable structures and physically meaningful temporal relations.

\paragraph{Limitations.}
This work has three main limitations. First, the structural vocabulary is discovered separately from encoder pretraining; although it can be iteratively refined, it is not jointly optimized with the encoder end to end. Second, while waveform-level analysis reveals interpretable atomic and emergent structures, their relationships to specific modulation components and physical mechanisms remain to be further investigated. Third, the atomic and emergent granularities are empirically selected and may not be optimal across different EM tasks or acquisition regimes. Future work will explore end-to-end vocabulary learning, deeper physical interpretation of structural units, and adaptive multi-granularity structural prediction.




\begin{ack}
This work was initiated and supported by the AI9Stars Team.

The authors declare no competing interests.
\end{ack}

\bibliographystyle{plainnat}
\bibliography{references}

\begin{thebibliography}{19}
\providecommand{\natexlab}[1]{#1}
\providecommand{\url}[1]{\texttt{#1}}
\expandafter\ifx\csname urlstyle\endcsname\relax
  \providecommand{\doi}[1]{doi: #1}\else
  \providecommand{\doi}{doi: \begingroup \urlstyle{rm}\Url}\fi

\bibitem[Baevski et~al.(2020)Baevski, Zhou, Mohamed, and Auli]{baevski2020wav2vec}
Alexei Baevski, Yuhao Zhou, Abdelrahman Mohamed, and Michael Auli.
\newblock wav2vec 2.0: A framework for self-supervised learning of speech representations.
\newblock \emph{Advances in neural information processing systems}, 33:\penalty0 12449--12460, 2020.

\bibitem[Chen et~al.(2025)Chen, Zu, Feng, Yang, and Li]{chen2025radiollm}
Shuai Chen, Yong Zu, Zhixi Feng, Shuyuan Yang, and Mengchang Li.
\newblock Radiollm: Introducing large language model into cognitive radio via hybrid prompt and token reprogrammings.
\newblock \emph{arXiv preprint arXiv:2501.17888}, 2025.

\bibitem[Devlin et~al.(2019)Devlin, Chang, Lee, and Toutanova]{devlin2019bert}
Jacob Devlin, Ming-Wei Chang, Kenton Lee, and Kristina Toutanova.
\newblock Bert: Pre-training of deep bidirectional transformers for language understanding.
\newblock In \emph{Proceedings of the 2019 conference of the North American chapter of the association for computational linguistics: human language technologies, volume 1 (long and short papers)}, pages 4171--4186, 2019.

\bibitem[Gage(1994)]{gage1994new}
Philip Gage.
\newblock A new algorithm for data compression.
\newblock \emph{The C Users Journal}, 12\penalty0 (2):\penalty0 23--38, 1994.

\bibitem[He et~al.(2016)He, Zhang, Ren, and Sun]{he2016deep}
Kaiming He, Xiangyu Zhang, Shaoqing Ren, and Jian Sun.
\newblock Deep residual learning for image recognition.
\newblock In \emph{Proceedings of the IEEE conference on computer vision and pattern recognition}, pages 770--778, 2016.

\bibitem[He et~al.(2022)He, Chen, Xie, Li, Doll{\'a}r, and Girshick]{he2022masked}
Kaiming He, Xinlei Chen, Saining Xie, Yanghao Li, Piotr Doll{\'a}r, and Ross Girshick.
\newblock Masked autoencoders are scalable vision learners.
\newblock In \emph{Proceedings of the IEEE/CVF conference on computer vision and pattern recognition}, pages 16000--16009, 2022.

\bibitem[Hsu et~al.(2021)Hsu, Bolte, Tsai, Lakhotia, Salakhutdinov, and Mohamed]{hsu2021hubert}
Wei-Ning Hsu, Benjamin Bolte, Yao-Hung~Hubert Tsai, Kushal Lakhotia, Ruslan Salakhutdinov, and Abdelrahman Mohamed.
\newblock Hubert: Self-supervised speech representation learning by masked prediction of hidden units.
\newblock \emph{IEEE/ACM transactions on audio, speech, and language processing}, 29:\penalty0 3451--3460, 2021.

\bibitem[Huang et~al.(2023)Huang, Pemasiri, Denman, Fookes, and Martin]{huang2023multi}
Zi~Huang, Akila Pemasiri, Simon Denman, Clinton Fookes, and Terrence Martin.
\newblock Multi-task learning for radar signal characterisation.
\newblock In \emph{2023 IEEE International Conference on Acoustics, Speech, and Signal Processing Workshops (ICASSPW)}, pages 1--5. IEEE, 2023.

\bibitem[Li et~al.(2025)Li, Deng, Xu, Feng, Schlegel, Huang, Sun, Sun, Yang, Yu, et~al.]{li2025mira}
Hao Li, Bowen Deng, Chang Xu, Zhiyuan Feng, Viktor Schlegel, Yu-Hao Huang, Yizheng Sun, Jingyuan Sun, Kailai Yang, Yiyao Yu, et~al.
\newblock Mira: Medical time series foundation model for real-world health data.
\newblock \emph{arXiv preprint arXiv:2506.07584}, 2025.

\bibitem[Luo et~al.(2025)Luo, Gui, Liu, Zhang, Zhang, Wang, Guo, Ma, Liu, He, et~al.]{luo2025emind}
Luqing Luo, Wenjin Gui, Yunfei Liu, Ziyue Zhang, Yunxi Zhang, Fengxiang Wang, Zonghao Guo, Zizhi Ma, Xinzhu Liu, Hanxiang He, et~al.
\newblock Emind: A foundation model for multi-task electromagnetic signals understanding.
\newblock \emph{arXiv preprint arXiv:2508.18785}, 2025.

\bibitem[Nie et~al.(2022)Nie, Nguyen, Sinthong, and Kalagnanam]{nie2022time}
Yuqi Nie, Nam~H Nguyen, Phanwadee Sinthong, and Jayant Kalagnanam.
\newblock A time series is worth 64 words: Long-term forecasting with transformers.
\newblock \emph{arXiv preprint arXiv:2211.14730}, 2022.

\bibitem[Oord et~al.(2018)Oord, Li, and Vinyals]{oord2018representation}
Aaron van~den Oord, Yazhe Li, and Oriol Vinyals.
\newblock Representation learning with contrastive predictive coding.
\newblock \emph{arXiv preprint arXiv:1807.03748}, 2018.

\bibitem[O'shea and West(2016)]{o2016radio}
Timothy~J O'shea and Nathan West.
\newblock Radio machine learning dataset generation with gnu radio.
\newblock In \emph{Proceedings of the GNU radio conference}, volume~1, 2016.

\bibitem[O’Shea et~al.(2018)O’Shea, Roy, and Clancy]{o2018over}
Timothy~James O’Shea, Tamoghna Roy, and T~Charles Clancy.
\newblock Over-the-air deep learning based radio signal classification.
\newblock \emph{IEEE Journal of Selected Topics in Signal Processing}, 12\penalty0 (1):\penalty0 168--179, 2018.

\bibitem[Scholl(2019)]{scholl2019classification}
Stefan Scholl.
\newblock Classification of radio signals and hf transmission modes with deep learning.
\newblock \emph{arXiv preprint arXiv:1906.04459}, 2019.

\bibitem[Su et~al.(2024)Su, Ahmed, Lu, Pan, Bo, and Liu]{su2024roformer}
Jianlin Su, Murtadha Ahmed, Yu~Lu, Shengfeng Pan, Wen Bo, and Yunfeng Liu.
\newblock Roformer: Enhanced transformer with rotary position embedding.
\newblock \emph{Neurocomputing}, 568:\penalty0 127063, 2024.

\bibitem[Vaswani et~al.(2017)Vaswani, Shazeer, Parmar, Uszkoreit, Jones, Gomez, Kaiser, and Polosukhin]{vaswani2017attention}
Ashish Vaswani, Noam Shazeer, Niki Parmar, Jakob Uszkoreit, Llion Jones, Aidan~N Gomez, {\L}ukasz Kaiser, and Illia Polosukhin.
\newblock Attention is all you need.
\newblock \emph{Advances in neural information processing systems}, 30, 2017.

\bibitem[Woo et~al.(2024)Woo, Liu, Kumar, Xiong, Savarese, and Sahoo]{woo2024unified}
Gerald Woo, Chenghao Liu, Akshat Kumar, Caiming Xiong, Silvio Savarese, and Doyen Sahoo.
\newblock Unified training of universal time series forecasting transformers.
\newblock In \emph{Forty-first International Conference on Machine Learning}, 2024.

\bibitem[Zhou et~al.(2025)Zhou, Liu, Zhang, Wu, Wu, Quek, and Chae]{zhou2025spectrumfm}
Fuhui Zhou, Chunyu Liu, Hao Zhang, Wei Wu, Qihui Wu, Tony~QS Quek, and Chan-Byoung Chae.
\newblock Spectrumfm: A foundation model for intelligent spectrum management.
\newblock \emph{IEEE Journal on Selected Areas in Communications}, 2025.

\end{thebibliography}








\clearpage
\appendix

\section{Pre-analysis Experiment}
\label{app:pre_analysis}

\begin{figure*}[h]
    \centering
    \begin{subfigure}[t]{0.49\textwidth}
        \centering
        \includegraphics[width=\linewidth]{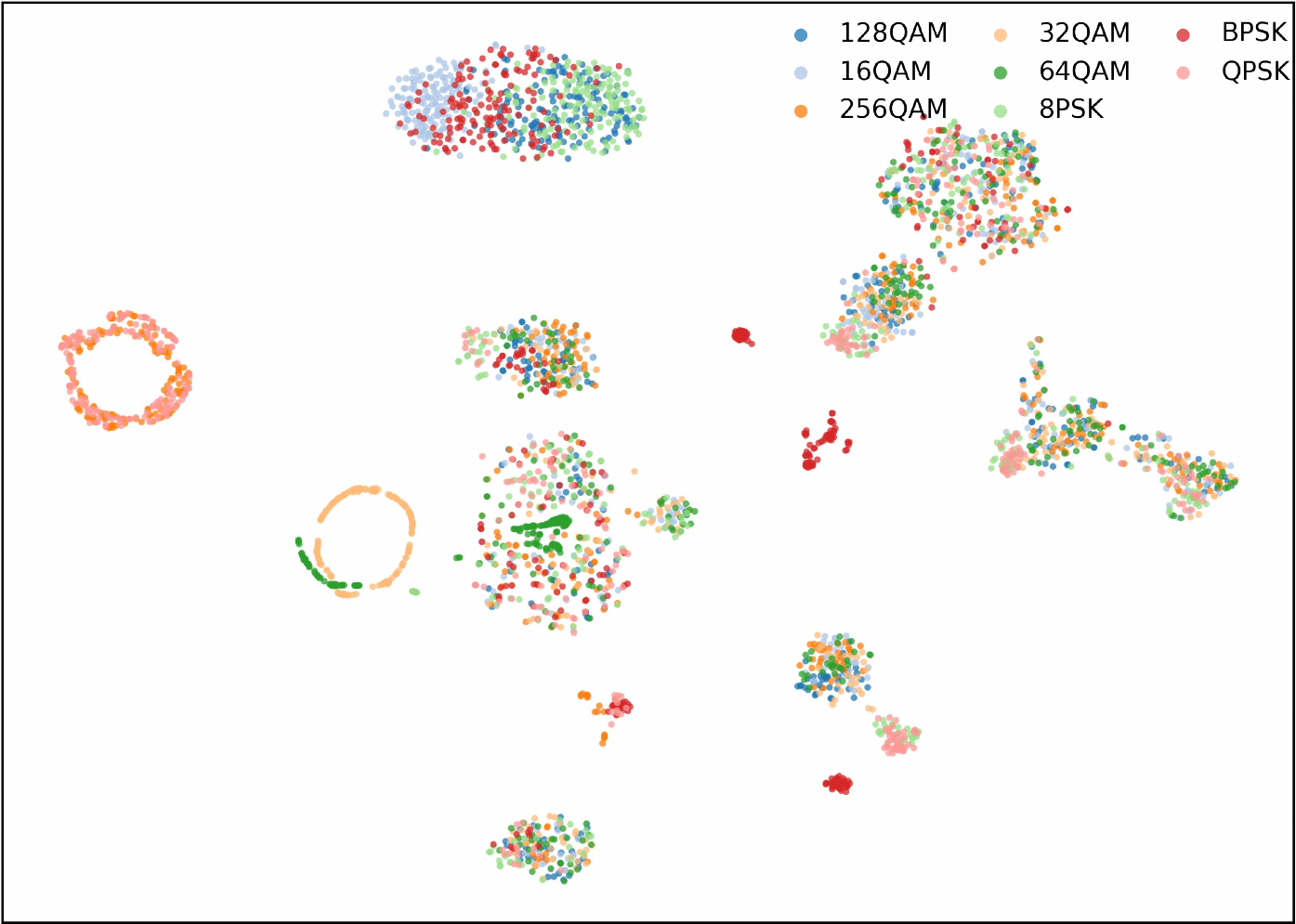}
        \caption{Cross-dataset heterogeneity, colored by modulation class.}
        \label{fig:vis_datasets_class}
    \end{subfigure}
    \hfill
    \begin{subfigure}[t]{0.49\textwidth}
        \centering
        \includegraphics[width=\linewidth]{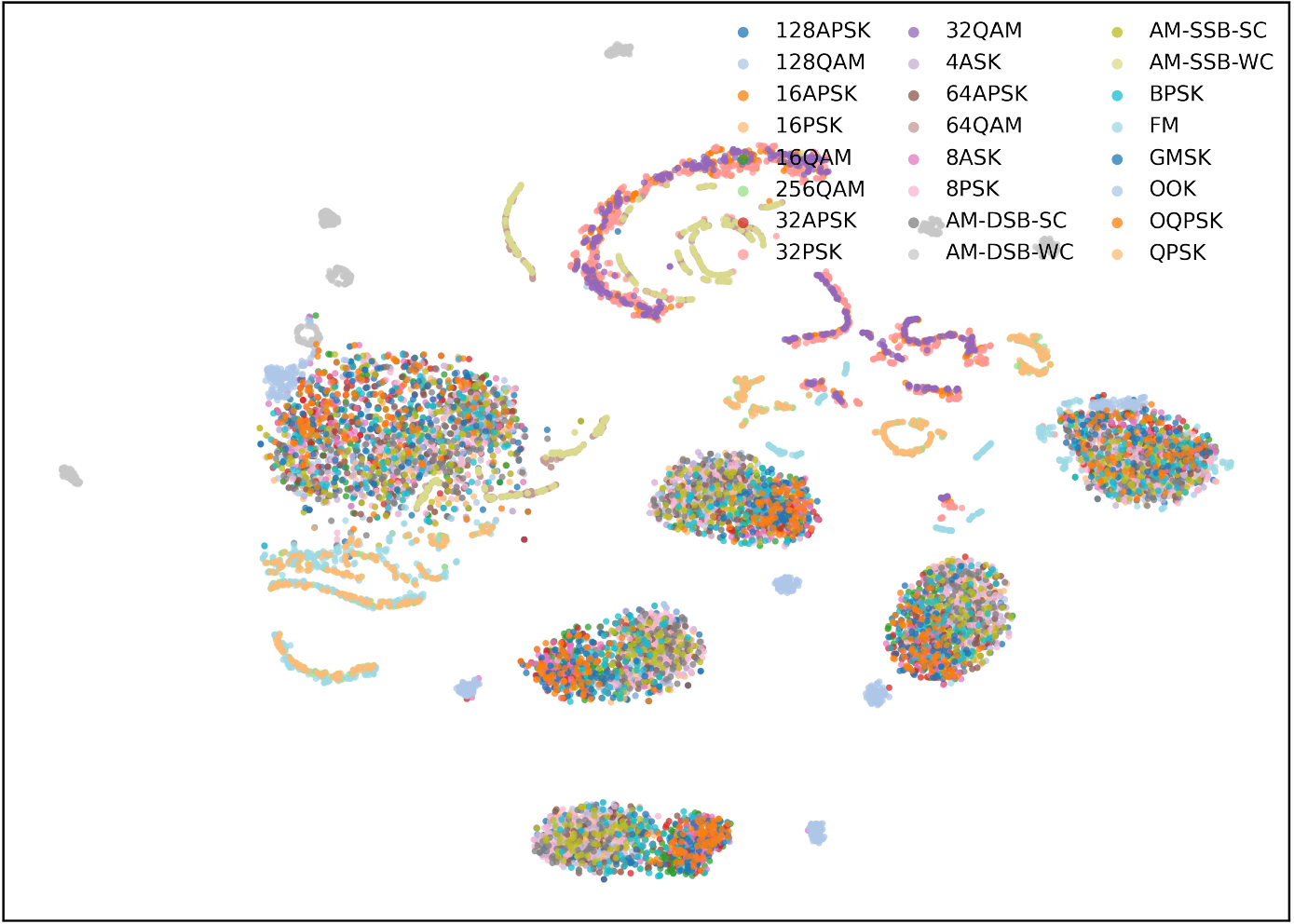}
        \caption{Sampling-rate heterogeneity, colored by modulation class.}
        \label{fig:vis_rml2018_class}
    \end{subfigure}

    \caption{
        Representation-space visualizations colored by modulation class, complementing the observation-conditioned visualizations in Figure~\ref{fig:heterogeneity_visualization}.
    }
    \label{fig:heterogeneity_visualization_appendix}
\end{figure*}

We conduct a pre-analysis experiment to examine whether observation conditions dominate the representation space of reconstruction-pretrained EM encoders. We consider two types of observation conditions: sampling-rate variation, which induces temporal-scale heterogeneity, and dataset-source variation, which reflects cross-dataset observational heterogeneity. The cosine-similarity analysis consists of three steps. 


First, we construct controlled sample sets for both sampling-rate and cross-dataset heterogeneity. For the sampling-rate experiment, we build multi-rate views from RML2018 (24 modulation classes, SNR = 18 dB). We randomly sample 150 anchor signals per class and generate resampled views using relative sampling-rate factors of $\{0.125,0.25,0.5,1,2,4,8\}$, yielding 25,200 representations. This produces a controlled multi-view set where the semantic class is preserved while the sampling-rate condition varies. For the cross-dataset experiment, we select eight shared modulation classes from RML2018, HIS2019, and EM-Comm at 18 dB SNR, sampling 150 signals per dataset and class (3,600 samples in total). We treat the dataset source as the observation condition, yielding another controlled sample set where modulation semantics are aligned, while acquisition settings, channel conditions, noise patterns, and data-generation processes may differ across dataset sources.

Second, we obtain pretrained representations for all samples. In the sampling-rate experiment, different resampled views share the same semantic label but correspond to different sampling-rate conditions. In the cross-dataset experiment, samples from different datasets are aligned by shared modulation classes while retaining their dataset-source labels. This allows us to compare how semantic classes and observation conditions organize the representation space. For visualization, we jointly project the representations using t-SNE with perplexity 40, 3,000 iterations, PCA initialization, and automatic learning rate. For the sampling-rate visualization, 11,928 balanced representations are selected from the 25,200 extracted representations.

Finally, we construct two groups of sample pairs in the representation space and compute cosine similarities. For the sampling-rate experiment, group A contains pairs from the same class but different sampling-rate factors and different anchors, measuring cross-rate intra-class consistency; group B contains pairs from different classes under the same sampling-rate factor, measuring within-rate inter-class similarity. For the cross-dataset experiment, group A contains pairs from the same class but different dataset sources, measuring cross-dataset intra-class consistency; group B contains pairs from different classes within the same dataset source, measuring within-dataset inter-class similarity. We report the mean similarities of the two groups as $\mu_A$ and $\mu_B$. For SpectrumFM, we use the only publicly available pretrained checkpoint, trained on RML2018, rather than the multi-dataset pretrained model reported in the original work.

The key diagnostic is the relationship between \(\mu_A\) and \(\mu_B\). If \(\mu_A < \mu_B\), samples from the same semantic class become less similar across observation conditions than samples from different classes under the same observation condition. This indicates that the representation space is more strongly organized by sampling rate or dataset source than by stable modulation structure. In other words, observation conditions can dominate semantic structure in the learned representation space, revealing an observation-biased representation induced by reconstruction-oriented pretraining.

\begin{table*}[t]
    \centering
    \caption{Cross-model cosine similarity analysis under observational heterogeneity. $A$ denotes same-structure pairs across different observations, and $B$ denotes different-structure pairs within the same observation. Cosine similarities are multiplied by 100, with $\Delta=A-B$. Results are reported as mean $\pm$ standard deviation over three runs.}
    \label{tab:observation_similarity}
    \small
    \setlength{\tabcolsep}{2pt}
    \begin{tabular}{llccccc}
        \toprule
        Model & Condition & $A$ & $B$ & $\Delta=A-B$ & Cohen's $d$ & $p$ \\
        \midrule
        EMind & Cross-dataset
        & $13.32 \pm 0.05$ & $32.29 \pm 0.37$
        & $-18.97 \pm 0.42$ & $-0.889 \pm 0.015$ & $<0.001$ \\

        EMind & Cross-sampling-rate
        & $19.78 \pm 0.13$ & $37.22 \pm 0.07$
        & $-17.44 \pm 0.06$ & $-0.681 \pm 0.003$ & $<0.001$ \\

        SpectrumFM & Cross-dataset
        & $84.53 \pm 0.01$ & $85.82 \pm 0.02$
        & $-1.29 \pm 0.02$ & $-0.219 \pm 0.002$ & $<0.001$ \\

        SpectrumFM & Cross-sampling-rate
        & $79.58 \pm 0.02$ & $85.61 \pm 0.02$
        & $-6.03 \pm 0.01$ & $-0.727 \pm 0.001$ & $<0.001$ \\
        \bottomrule
    \end{tabular}
\end{table*}

As shown in Table~\ref{tab:observation_similarity}, both EMind and SpectrumFM exhibit lower cross-observation intra-class similarity ($A$) than within-observation inter-class similarity ($B$) under both dataset-source and sampling-rate variation. The consistently negative $\Delta$ values provide quantitative evidence of observation bias across reconstruction-pretrained encoders.

We further assess the observation information retained in pretrained representations through linear probing. With frozen encoders, we train linear classifiers on the validation splits of RML2018, EM-Comm, and HIS2019 and evaluate them on the corresponding test splits. Both dataset-source and modulation classification are evaluated as three-class tasks for a matched comparison.

\begin{table}[t]
    \centering
    \caption{Linear-probing accuracy (\%) for dataset-source and modulation classification using frozen pretrained representations. Both tasks contain three classes.}
    \label{tab:observation_linear_probe}
    \small
    \setlength{\tabcolsep}{5pt}
    \begin{tabular}{lccc}
        \toprule
        Model & Dataset & Modulation & Chance \\
        \midrule
        EMind & 99.22 & 62.14 & 33.33 \\
        SpectrumFM & 79.06 & 57.89 & 33.33 \\
        \bottomrule
    \end{tabular}
\end{table}

As shown in Table~\ref{tab:observation_linear_probe}, both models achieve substantially above-chance accuracy in dataset-source classification, exceeding their modulation classification accuracy. These results indicate that observation-specific information remains readily accessible in pretrained representations, complementing the cosine-similarity analysis.

\section{Experiment Setup}
\textbf{Atomic dictionary construction.} During atomic dictionary construction, we adopt balanced sampling across datasets. Each signal is normalized by the 99th percentile of absolute amplitudes and partitioned into non-overlapping patches with patch size \(ps=8\). In Iter0, K-Means clustering is performed directly on patch-level raw features. In Iter1, we reload the Iter0-pretrained encoder, extract token-level representations, and perform K-Means clustering on these refined features. The number of atomic clusters is set to \(K_a=512\), with a mini-batch size of 16,384 and 5 independent initializations. The resulting atomic dictionary maps each I/Q sequence into a discrete token sequence.

\textbf{Emergent dictionary construction.} We apply BPE merging on the atomic token sequences using 20M training tokens. A total of $K_e=512$ merge operations are performed with a minimum frequency threshold of 20,
and the maximum emergent unit length is capped at 5 atoms. 
The full ESU vocabulary used for emergent prediction includes both the original $K_a=512$ atomic singleton units and the $K_e=512$ merged units, resulting in $K_a+K_e=1024$ prediction targets.

\textbf{Structural pretraining.} In each iteration, we normalize the original signal and partition it into non-overlapping patches. We apply span masking with a masking probability of \(0.06\) and a mask span length of \(8\). The model is then trained with cross-entropy loss to predict the discrete structural-unit label corresponding to each masked position.

\textbf{Iterative refinement.} Following HuBERT-style refinement, we perform two rounds of dictionary construction and structural pretraining.
In Iter0, the atomic dictionary is constructed directly from local patch-level raw features, and the emergent vocabulary is induced from the resulting atomic label sequences.
After the first pretraining round, Iter1 reloads the Iter0-pretrained encoder, re-extracts token-level representations, rebuilds the atomic and emergent dictionaries, and conducts a second pretraining round.

\textbf{Reference sampling rate in RoPE-fs.}
We set the reference sampling rate \(f_0\) to 1 MHz for all experiments. Given the sampling rate \(f_i\) of each input signal, RoPE-fs maps discrete patch indices to normalized physical-time positions through the ratio \(f_0/f_i\). Therefore, \(f_0\) serves as a fixed unit reference rather than a dataset-specific hyperparameter. We keep \(f_0\) unchanged across all datasets to ensure a consistent physical-time scale and reproducible positional encoding.

\section{Optimization details}

For pretraining, we use the AdamW optimizer with a linear warmup followed by cosine learning-rate decay. The model is pretrained for 100 epochs with a peak learning rate of \(3.6\times 10^{-3}\), weight decay of \(0.01\), warmup ratio of \(0.18\), and minimum learning rate of \(0\). For downstream fine-tuning, we also use AdamW with the same warmup-cosine scheduling strategy. By default, downstream models are trained for 40 epochs with batch size \(256\), learning rate \(1\times 10^{-4}\), and weight decay \(0.08\), unless otherwise specified. We use fixed random seeds to ensure reproducibility, with repeated-run statistics reported separately where applicable. Few-shot fine-tuning follows the same optimizer and scheduling configuration as full-data fine-tuning, differing only in the sampling strategy and the number of labeled training examples.

\section{Pretraining Data}
\label{app:pretrain_data}

We construct the unlabeled pretraining corpus from the dataset suite proposed by EMind~\citep{luo2025emind}. 
Rather than using the full available corpus, this work uses a dataset-balanced subset consisting of six datasets: RadarCommDataset, WiSig, HisarMod2019.1, EM-Comm, EM-Radar, and EM-Infer-Radar-v2. 
To prevent large-scale datasets from dominating pretraining, we cap the number of samples from each dataset at 300,000. 
For datasets with fewer than 300,000 available samples, we use all available samples. 
This results in approximately 1.62M pretraining sequences in total, as summarized in Table~\ref{tab:pretrain_subset}. 
All dictionary construction, emergent-unit mining, and self-supervised pretraining are performed only on this selected pretraining corpus.


\begin{table}[t]
\centering
\caption{Pretraining subset used in this work.}
\label{tab:pretrain_subset}
\small
\setlength{\tabcolsep}{4pt}
\begin{tabular}{lccc}
\toprule
\textbf{Dataset Name} & \textbf{Signal Length} & \textbf{Sampling Rate} & \textbf{Used Samples} \\
\midrule
EM-Comm              & 1024 & 20 MHz & 300,000 \\
HisarMod2019.1       & 1024 & 1 MHz  & 300,000 \\
RadarCommDataset     & 128  & 10 MHz & 300,000 \\
EM-Radar             & 1024 & 5 MHz  & 300,000 \\
WiSig                & 256  & 25 MHz & 300,000 \\
EM-Infer-Radar-v2    & 2048 & 20 MHz & 120,000 \\
\midrule
Total                & --   & --     & 1,620,000 \\
\bottomrule
\end{tabular}
\end{table}

\section{Finetuning Dataset}
For downstream evaluation, we use RML2016.10A, RML2016.04C, RML2016.10B, RML2018.01A~\citep{o2016radio, o2018over}, Panoradio~\citep{scholl2019classification}, RadChar~\citep{huang2023multi}, EM-Infer-Comm, EM-Infer-Radar, and EM-AIS~\citep{luo2025emind}, as shown in Table~\ref{tab:finetune_data}.

\begin{table}[t]
\centering
\caption{Fine-tuning datasets.}
\label{tab:finetune_data}
\small
\setlength{\tabcolsep}{3pt}
\begin{tabular}{lcccc}
\toprule
\textbf{Dataset} & \textbf{Length} & \textbf{Classes} & \textbf{SNR (dB)} & \textbf{Rate} \\
\midrule
RML2016.10A    & 128  & 11  & $-20\sim18$ & 1 MHz \\
RML2016.10B    & 128  & 10  & $-20\sim18$ & 1 MHz \\
RML2016.04C    & 128  & 11  & $-20\sim18$ & 1 MHz \\
RML2018.01A    & 1024 & 24  & $-20\sim20$ & 3.2 MHz \\
Panoradio HF   & 2048 & 18  & $-10\sim25$ & 172.8 kHz \\
RadChar        & 512  & 5   & $-20\sim20$ & 3.2 MHz \\
EM-Infer-Comm  & 1024 & 9   & $-20\sim20$ & 20 MHz \\
EM-Infer-Radar & 1024 & 12  & $-20\sim20$ & 20 MHz \\
EM-AIS         & 3840 & 112 & $-20\sim20$ & 156.25 MHz \\
\bottomrule
\end{tabular}
\end{table}

\section{Further Analyses}
\label{app:further_analyses}

\paragraph{Additional component ablations.}
We conduct additional controlled ablations on RML2016.10A and RadChar to isolate the contributions of multi-level structural supervision, iterative dictionary refinement, and RoPE-fs. As shown in Table~\ref{tab:additional_component_ablations}, multi-level supervision outperforms either atomic-only or emergent-only supervision. Iterative dictionary refinement also yields larger gains than compute-matched continued training with fixed dictionaries, while RoPE-fs consistently improves upon standard RoPE.

\begin{table}[t]
    \centering
    \caption{Additional component ablations on RML2016.10A and RadChar (accuracy, \%). Iter0 continued uses an additional 100 training epochs with fixed dictionaries.}
    \label{tab:additional_component_ablations}
    \small
    \setlength{\tabcolsep}{3pt}
    \begin{tabular}{llcc}
        \toprule
        Ablation & Setting & RML16A & RadChar \\
        \midrule
        Targets & Atomic-only & 62.78 & 89.11 \\
                & Emergent-only & 62.23 & 89.15 \\
                & Multi-level & \textbf{64.07} & \textbf{89.85} \\
        \midrule
        Iterations & Iter0 & 62.04 & 89.01 \\
                   & Iter0 continued & 62.22 & 89.29 \\
                   & Iter1 refinement & \textbf{64.07} & \textbf{89.85} \\
        \midrule
        Position & Standard RoPE & 61.86 & 89.28 \\
                 & RoPE-fs & \textbf{64.07} & \textbf{89.85} \\
        \bottomrule
    \end{tabular}
\end{table}

\paragraph{Hyperparameter sensitivity.}
We evaluate the sensitivity of six key hyperparameters on RML2016.10A and RadChar, varying one parameter at a time while keeping the others fixed. As shown in Table~\ref{tab:hyperparameter_sensitivity}, the adopted configuration achieves the highest accuracy among the evaluated settings for both datasets. Performance is particularly sensitive to larger patch sizes, while the results support the selected vocabulary sizes, ESU length, mask ratio, and reference sampling rate.

\begin{table}[t]
    \centering
    \caption{Hyperparameter sensitivity on RML2016.10A and RadChar (accuracy, \%). Results correspond to the settings in the listed order, with the adopted configurations highlighted in bold.}
    \label{tab:hyperparameter_sensitivity}
    \small
    \setlength{\tabcolsep}{5pt}
    \begin{tabular}{llcc}
        \toprule
        Hyperparameter & Settings & RML2016.10A & RadChar \\
        \midrule
        Reference rate $f_0$ & 0.5 / 1 / 2 MHz
        & 62.01 / \textbf{64.07} / 63.51
        & 88.82 / \textbf{89.85} / 89.76 \\

        Patch size $P$ & 4 / 8 / 16
        & 61.91 / \textbf{64.07} / 56.56
        & 89.57 / \textbf{89.85} / 88.60 \\

        ASU vocabulary size $K_a$ & 256 / 512 / 1024
        & 61.98 / \textbf{64.07} / 62.20
        & 88.95 / \textbf{89.85} / 89.11 \\

        ESU merge count $K_e$ & 256 / 512 / 1024
        & 62.62 / \textbf{64.07} / 62.58
        & 89.17 / \textbf{89.85} / 88.92 \\

        Maximum ESU length & 3 / 5 / 8
        & 62.93 / \textbf{64.07} / 62.81
        & 89.69 / \textbf{89.85} / 89.68 \\

        Mask ratio & 0.03 / 0.06 / 0.10
        & 63.23 / \textbf{64.07} / 62.09
        & 89.68 / \textbf{89.85} / 89.18 \\
        \bottomrule
    \end{tabular}
\end{table}

\paragraph{Factor-level sampling-rate extrapolation.}

\begin{figure}[t]
    \centering
    \includegraphics[width=\columnwidth]{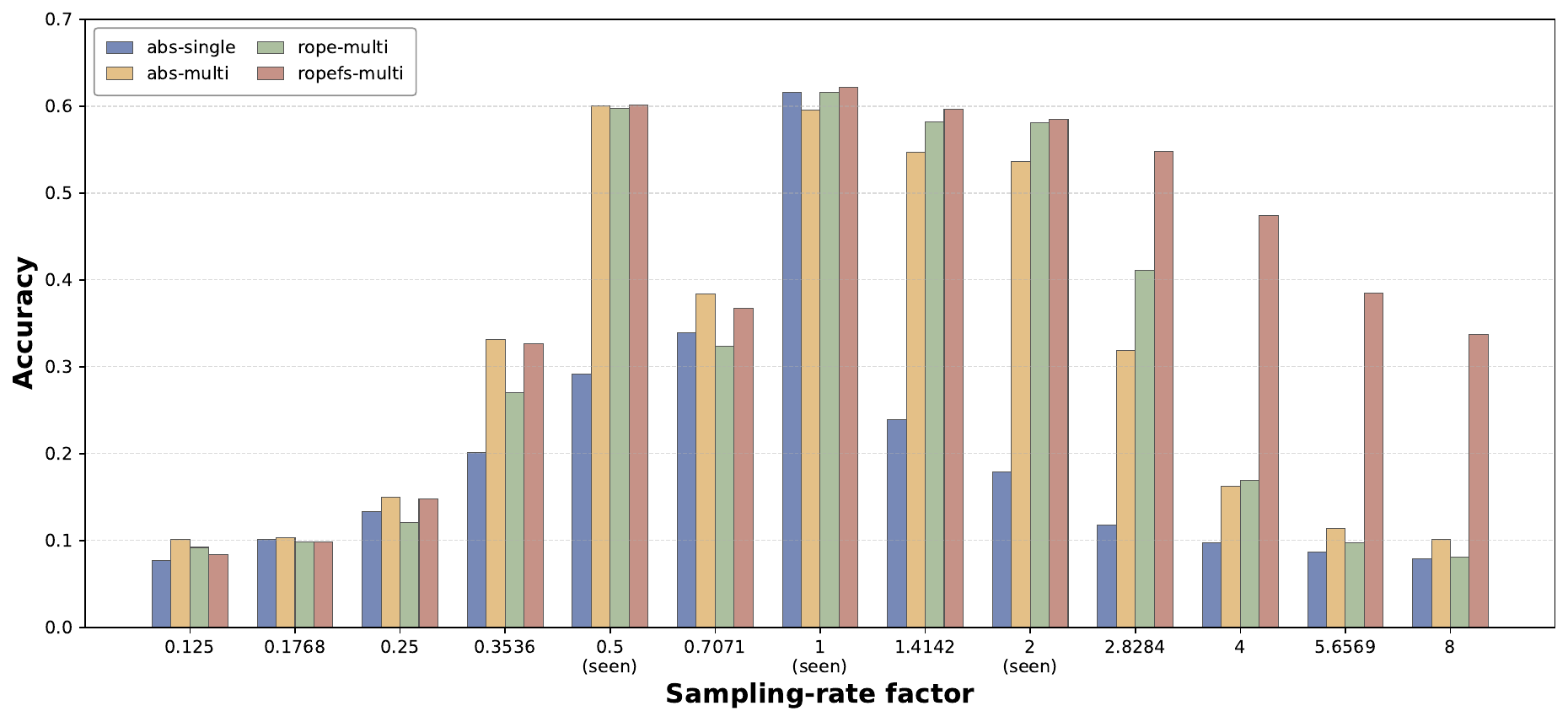}
    \caption{Factor-level sampling-rate extrapolation results on RML18A.}
    \label{fig:sampling_rate_factor}
\end{figure}

Figure~\ref{fig:sampling_rate_factor} provides a factor-level view of the sampling-rate extrapolation experiment on RML18A. The results complement the aggregate seen/unseen comparison in the main text by showing how different positional designs behave across individual resampling factors. Absolute positional encoding trained on a single rate suffers clear degradation once the test factor moves away from the seen rates. Multi-rate training improves robustness to some extent, but absolute and standard RoPE encodings still show unstable extrapolation across factors. In contrast, RoPE-fs consistently achieves stronger performance on most unseen factors, especially under larger sampling-rate shifts. This supports our claim that aligning positional phases with physical-time coordinates improves temporal-scale generalization.

\paragraph{Sampling-rate metadata robustness.}
We evaluate the robustness of RoPE-fs to inaccurate and missing sampling-rate metadata on RML2016.10A and RadChar. For inaccurate metadata, we keep the input I/Q signals unchanged and perturb the sampling rates provided to RoPE-fs by $\pm 5\%$ and $\pm 10\%$ at inference time. For missing metadata, we use the reference sampling rate $f_0=1$ MHz during both fine-tuning and inference to maintain consistent positional encoding.

\begin{table}[t]
    \centering
    \caption{Robustness to inaccurate and missing sampling-rate metadata (accuracy, \%). Sampling-rate errors are introduced at inference time, while missing metadata is handled using $f_0$ during both fine-tuning and inference.}
    \label{tab:sampling_rate_metadata_robustness}
    \small
    \setlength{\tabcolsep}{3pt}
    \begin{tabular}{lcccccc}
        \toprule
        Dataset & Standard & $-5\%$ & $+5\%$ & $-10\%$ & $+10\%$ & Missing \\
        \midrule
        RML2016.10A & 64.07 & 63.13 & 63.04 & 63.18 & 62.91 & 64.07 \\
        RadChar     & 89.85 & 89.59 & 89.36 & 89.00 & 88.72 & 89.24 \\
        \bottomrule
    \end{tabular}
\end{table}

As shown in Table~\ref{tab:sampling_rate_metadata_robustness}, RoPE-fs exhibits limited performance degradation under sampling-rate errors of up to 10\%. For missing metadata, the fallback strategy preserves performance on both datasets. In particular, RML2016.10A has the same sampling rate as $f_0$, making the fallback equivalent to the standard setting, while RadChar retains 89.24\% accuracy despite its different sampling rate. These results demonstrate the robustness of RoPE-fs to moderate sampling-rate inaccuracies and missing metadata.

\paragraph{Dictionary quality analysis.}

\begin{table*}[t]
\centering
\caption{Dictionary quality analysis.}
\label{tab:dict_quality}

\begin{subtable}[t]{0.46\textwidth}
\centering
\caption{Atomic dictionary}
\label{tab:atomic_dict_quality}
\resizebox{\linewidth}{!}{%
\begin{tabular}{llc}
\toprule
\textbf{Aspect} & \textbf{Metric} & \textbf{Value} \\
\midrule
Usage & Used token ratio $\uparrow$ & 66.38 \\
Usage & Normalized entropy $\uparrow$ & 79.69 \\
Usage & Top-5 usage ratio $\downarrow$ & 12.20 \\
\midrule
Structure & Per-sample entropy & 39.23 \\
Structure & Bigram entropy & 55.53 \\
\midrule
Dependency & NMI w/ mode & 13.51 \\
Dependency & NMI w/ dataset & 12.95 \\
\bottomrule
\end{tabular}%
}
\end{subtable}
\hspace{0.015\textwidth}
\begin{subtable}[t]{0.48\textwidth}
\centering
\caption{Emergent dictionary}
\label{tab:bpe_dict_quality}
\resizebox{\linewidth}{!}{%
\begin{tabular}{llc}
\toprule
\textbf{Aspect} & \textbf{Metric} & \textbf{Value} \\
\midrule
Usage & Used vocab. ratio $\uparrow$ & 99.90 \\
Usage & Token entropy $\uparrow$ & 92.37 \\
\midrule
Compositionality & Avg. token length $\uparrow$ & 1.55 \\
Compositionality & Compression ratio $\downarrow$ & 64.55 \\
Compositionality & Long-unit usage ratio $\uparrow$ & 24.84 \\
\bottomrule
\end{tabular}%
}
\end{subtable}
\end{table*}

Table~\ref{tab:dict_quality} summarizes the quality of the learned atomic and emergent dictionaries. For the atomic dictionary, the used token ratio and normalized entropy indicate that the codebook is not collapsed to a small number of frequent tokens, while the low top-5 usage ratio further suggests balanced token utilization. The non-trivial per-sample and bigram entropies show that atomic tokens form diverse temporal patterns rather than nearly constant assignments. Meanwhile, the moderate NMI values with modulation mode and dataset source indicate that the atomic units are not simply global class or dataset identifiers, but provide local structural targets for masked prediction.

For the emergent dictionary, the nearly full vocabulary usage and high token entropy show that BPE merging produces an active and well-utilized higher-level vocabulary. The average token length, compression ratio, and long-unit usage ratio further indicate that the emergent units capture recurring variable-length compositions beyond isolated patch-level atoms. Together, these results suggest that the learned dictionaries provide compact yet diverse structural supervision, supporting the effectiveness of masked multi-level structural prediction.

We further examine the sequential organization of structural units using 15,000 validation signals from RML2018, HIS2019, and EM-Comm. The marginal entropy $H(T)$ is 5.04 nats, which decreases to a conditional entropy $H(T_{t+1}\mid T_t)$ of 3.06 nats when the preceding token is known, corresponding to a 39.3\% reduction in uncertainty. This demonstrates that the learned structural units exhibit non-random temporal dependencies and organized sequential patterns.

\paragraph{Structural unit stability under perturbations.}
We evaluate the stability of learned structural units under common signal perturbations on HIS2019 and EM-Comm. We construct paired views of the same signal using additive white Gaussian noise (AWGN; 5/10/20 dB), phase rotations (30/60/90 degrees), normalized carrier-frequency offsets (CFO; 0.002/0.005/0.01), and temporal cyclic shifts (3/5/10\% of sequence length). We compare two pipelines: directly clustering raw I/Q patches and clustering representations extracted by the pretrained encoder. Stability is measured by cosine similarity for continuous representations and token matching rate (TMR) for discrete structural units.

\begin{table}[!htbp]
    \centering
    \caption{Stability of atomic and emergent structural units under signal perturbations. Each entry compares raw I/Q-based clustering with pretrained encoder-based clustering (Raw $\rightarrow$ Pretrained). TMR is reported in \%, and cosine similarities are multiplied by 100. Higher values indicate greater consistency between paired views.}
    \label{tab:structural_unit_stability}
    \small
    \setlength{\tabcolsep}{6pt}
    \begin{tabular}{lcccc}
        \toprule
        Perturbation & TMR ASU $\uparrow$ & TMR ESU $\uparrow$ & Cosine ASU $\uparrow$ & Cosine ESU $\uparrow$ \\
        \midrule
        AWGN           & 62.9 $\rightarrow$ 64.2 & 67.8 $\rightarrow$ 69.9 & 91.8 $\rightarrow$ 91.7 & 93.0 $\rightarrow$ 89.6 \\
        Phase rotation & 5.7 $\rightarrow$ 19.3  & 4.4 $\rightarrow$ 12.4  & 45.5 $\rightarrow$ 74.1 & 45.4 $\rightarrow$ 62.3 \\
        CFO            & 9.6 $\rightarrow$ 15.3  & 8.6 $\rightarrow$ 9.7   & 2.0 $\rightarrow$ 67.1  & 1.0 $\rightarrow$ 54.9 \\
        Cyclic shift   & 2.5 $\rightarrow$ 8.8   & 4.3 $\rightarrow$ 10.2  & -0.8 $\rightarrow$ 63.0 & -0.4 $\rightarrow$ 53.7 \\
        \midrule
        Overall        & \textbf{20.2 $\rightarrow$ 26.9}
                       & \textbf{21.3 $\rightarrow$ 25.6}
                       & \textbf{34.6 $\rightarrow$ 74.0}
                       & \textbf{34.8 $\rightarrow$ 65.1} \\
        \bottomrule
    \end{tabular}
\end{table}

As shown in Table~\ref{tab:structural_unit_stability}, pretrained encoder representations exhibit substantially higher cosine similarity under phase rotations, frequency offsets, and temporal shifts, while maintaining high similarity under AWGN. The token matching rates of both ASUs and ESUs also improve across all perturbation types. These results indicate that pretraining enhances cross-perturbation consistency in continuous representations, which is further reflected in more stable discrete structural-unit assignments.

\FloatBarrier

\section{Compute Resources}

All experiments are implemented in PyTorch and conducted on CUDA-enabled NVIDIA GPU platforms. Large-scale pretraining is performed on NVIDIA A800 GPUs with distributed data parallel training, using up to 8 GPUs per node with \texttt{nproc\_per\_node=8}. To accommodate variable-length electromagnetic sequences, we control the effective batch size by a token budget, with \texttt{tokens\_per\_batch=81920}.

For large-scale pretraining, one training round on 8 A800 GPUs takes approximately 5 hours. The exact runtime depends on the dataset mixture, sequence-length distribution, and data-loading throughput. For memory efficiency, large-scale training uses a streaming data-loading mode, where parquet files are read in chunks with \texttt{data\_mode=stream}. Smaller ablation and evaluation experiments use in-memory loading when feasible.

Downstream full-supervision fine-tuning and few-shot fine-tuning are run on either single-GPU or multi-GPU settings depending on the scale of each benchmark. Their runtime varies substantially across tasks because the downstream datasets differ in both the number of samples and the input sequence length. In general, few-shot fine-tuning and small-scale ablations require substantially less compute than large-scale pretraining, while full-supervision fine-tuning on larger datasets requires more runtime.

To improve reproducibility, we fix random seeds for all experiments and log the complete training configuration, including the optimizer, learning-rate schedule, masking parameters, data-loading mode, and checkpoint paths.

\section{Broader impacts}

This work may benefit spectrum monitoring, wireless communication, radar sensing, and robust electromagnetic signal analysis. Potential negative uses include unauthorized signal interception, emitter identification, or surveillance-related applications. We encourage responsible use only in authorized sensing and communication scenarios, and any deployment should follow applicable regulations, privacy constraints, and spectrum-management policies.



\end{document}